\documentclass[runningheads]{llncs}

\usepackage[dvipsnames]{xcolor}

\newcommand{\TODO}[1]{\textbf{\color{red}TODO}}

\definecolor{cvprblue}{rgb}{0.21,0.49,0.74}
\usepackage[breaklinks,colorlinks,citecolor=cvprblue,linkcolor=cvprblue,urlcolor=cvprblue]{hyperref}

\usepackage{fontawesome5}

\usepackage{amsmath,amssymb}
\usepackage{url}
\usepackage[T1]{fontenc}
\usepackage{pifont}
\usepackage{graphicx}
\usepackage{glossaries}
\glsdisablehyper
\usepackage{pgfplots}
\usepackage{pgfplotstable}
\usepackage{multirow}
\usepackage{tikz}
\usetikzlibrary{shadows,shadows.blur,patterns}
\usepackage{cuted}
\usepackage{ragged2e}
\usepackage{algorithm}
\usepackage[noend]{algpseudocode}
\usepackage[shortlabels]{enumitem} 
\usepackage[capitalize]{cleveref}
\usepackage[numbers,sort&compress]{natbib}
\makeatletter
\renewcommand{\@biblabel}[1]{#1.}
\makeatother
\usepackage{subcaption}
\usepackage{helvet}
\usepackage[eulergreek]{sansmath}

\tikzfading[name=fade down,
    top color=transparent!0,
    bottom color=transparent!20]

\usepgfplotslibrary{fillbetween,colorbrewer}

\pgfplotsset{
  log ticks with fixed point,
}

\pgfdeclareplotmark{01}{\pgfuseplotmark{+}}
\pgfdeclareplotmark{02}{\pgfuseplotmark{triangle}}
\pgfdeclareplotmark{03}{\pgfuseplotmark{square}}

\pgfplotscreateplotcyclelist{scatter}{%
    {Aquamarine,fill=Aquamarine!100!white,draw=white, draw opacity = 0.4},
    {Orchid,fill=Orchid!100!white,draw=white, draw opacity = 0.4},
    {RoyalBlue,fill=RoyalBlue!100!white,draw=white, draw opacity = 0.4},
    {Cyan,fill=Cyan!100!white,draw=white, draw opacity = 0.4},
    {Melon,fill=Melon!100!white,draw=white, draw opacity = 0.4},
    {Goldenrod,fill=Goldenrod!100!white,draw=white, draw opacity = 0.4},
    {VioletRed,fill=VioletRed!100!white,draw=white, draw opacity = 0.4}
}

\pgfplotsset{
    /pgfplots/bar cycle list/.style={/pgfplots/cycle list={
        {Cyan,fill=Cyan!100!white,mark=none},
        {VioletRed,fill=VioletRed!100!white,mark=none},
        {RoyalBlue,fill=RoyalBlue!100!white,mark=none}
        },
    },
}

\pgfplotsset{
    colormap={confmatrix}{
        rgb255=(255, 255, 255);
        rgb255=(130, 245, 255);
        rgb255=(218, 165, 32);
        rgb255=(253, 188, 180);
        rgb255=(238, 130, 238);
        rgb255=(158, 52, 154);
    },
}

\makeatletter \newcommand{\pgfplotsdrawaxis}{\pgfplots@draw@axis} \makeatother

\pgfplotsset{axis line on top/.style={
  axis line style=transparent,
  ticklabel style=transparent,
  tick style=transparent,
  axis on top=false,
  after end axis/.append code={
    \pgfplotsset{
      axis line style=opaque,
      ticklabel style=opaque,
      tick style=opaque,
      grid=none
    }
    \pgfplotsdrawaxis
  }
  }
}

\pgfplotsset{every tick label/.append style={font=\small}}

\definecolor{bar_chart_1}{RGB}{90, 180, 180}
\definecolor{bar_chart_2}{RGB}{82, 109, 136}
\definecolor{bar_chart_3}{RGB}{120, 162, 227}

\makeatletter
\pretocmd{\NAT@citexnum}{\@ifnum{\NAT@ctype>\z@}{\let\NAT@hyper@\relax}{}}{}{}
\makeatother

\newcommand{\oursmajority}{81.01}

\newcommand{\oursmead}{79.34}

\newcommand{\cmark}{\ding{51}}%
\let\titleold\title
\renewcommand{\title}[1]{\titleold{#1}\newcommand{\thetitle}{#1}}
\def\maketitlesupplementary
   {
   \newpage
       \onecolumn
        \centering
        \Large
        \textbf{\thetitle}\\
        \vspace{0.5em}Supplementary Material \\
        \vspace{1.0em}
        \normalsize
        \justifying
   }

\def\onedot{.~}

\def\ie{\emph{i.e}\onedot}

\crefname{section}{Sec.}{Secs.}
\Crefname{section}{Section}{Sections}
\Crefname{table}{Table}{Tables}
\crefname{table}{Tab.}{Tabs.}

\newcommand{\confmatrix}[1]{
\begin{tikzpicture}
    \begin{axis}[
        width=\confmatsize\linewidth,
        height=\confmatsize\linewidth,
        colormap name=confmatrix,
        xlabel=Predicted,
        xlabel style={yshift=-5pt, font=\tiny\sansmath\sffamily},
        ylabel=True,
        ylabel style={yshift=-10pt, font=\tiny\sansmath\sffamily},
        xticklabels={calm, happy, sad, angry, fearful, disgust, surprised}, % changed
        xticklabel style={font=\tiny\sansmath\sffamily,rotate=45},
        xtick={0,...,6}, % changed
        xtick style={draw=none},
        yticklabels={calm, happy, sad, angry, fearful, disgust, surprised}, % changed
        yticklabel style={font=\tiny\sansmath\sffamily},
        ytick={0,...,6}, % changed
        ytick style={draw=none},
        enlargelimits=false,
        font=\tiny,
        axis line style={white},
        point meta min=0.0,
        point meta max=1.0,
        nodes near coords={\pgfmathprintnumber\pgfplotspointmeta},
        nodes near coords black white/.style={
            small value/.style={
                font=\tiny\sansmath\sffamily,
                yshift=-5pt,
                text=black,
                /pgf/number format/fixed,
                /pgf/number format/precision=2
            },
            large value/.style={
                font=\tiny\sansmath\sffamily,
                yshift=-5pt,
                text=white,
                /pgf/number format/fixed,
                /pgf/number format/precision=2
            },
            every node near coord/.style={
                check for zero/.code={
                    \begingroup
                        \pgfkeys{/pgf/fpu}
                        \pgfmathparse{\pgfplotspointmeta<0.3}
                        \global\let\resultsmall=\pgfmathresult
                    \endgroup
                    \pgfmathfloatcreate{1}{1.0}{0}
                    \let\ONE=\pgfmathresult
                    \ifx\resultsmall\ONE
                        \pgfkeysalso{/pgfplots/small value}
                    \else
                        \pgfkeysalso{/pgfplots/large value}
                    \fi
                },
                check for zero,
            },
        },
        nodes near coords black white
    ]
    \addplot[
        matrix plot,
        mesh/cols=7, % changed
        point meta=explicit
    ] table [meta=prob, col sep=comma] {#1}; % added every entry where x=4 or y=4
    \end{axis}
\end{tikzpicture}
}

\newacronym{fer}{FER}{facial expression recognition}
\newacronym{pog}{PoG}{product of gaussian distributions}
\newacronym{face_reconst_net}{FRN}{face reconstruction net}
\newacronym{context_reconst_net}{CRN}{context reconstruction net}
\newacronym{context_att_net}{CAN}{context attention net}
\newacronym{vaegan}{VAE-GAN}{variational-autoencoder general adversarial network}
\newacronym{vae}{VAE}{variational-autoencoder}
\newacronym{ae}{AE}{autoencoder}
\newacronym{gan}{GAN}{generative adversarial network}
\newacronym{kl_divergence}{KL-Divergence}{Kullback–Leibler divergence}
\newacronym{pregc}{PREG-C}{photo-realistic expression generation under context}
\newacronym{var_inf}{VI}{variational inference}
\newacronym{em}{EM}{expectation maximization}
\newacronym{poe}{PoE}{product of experts}
\newacronym{ann}{ANN}{artificial neural network}
\newacronym{ai}{AI}{artificial intelligence}
\newacronym{ci}{CI}{classification image}
\newacronym{sota}{SOTA}{state-of-the-art}

\newglossaryentry{overfitting}{%
name=overfitting,
description={a model that has learned to perform  on the training data very well, but performs poorly on unseen data is overfitting}
}
\newglossaryentry{mse}{
name=MSE,
description=Mean Squared Error
}
\newglossaryentry{rafd}{
name=RaFD,
description=Radboud faces database}
\newglossaryentry{ravdess}{
name=RAVDESS,
description=Ryerson audio-visual database of emotional speech and song}
\newglossaryentry{mead}{
name=MEAD,
description=MEAD dataset}
\newglossaryentry{celeb_db}{
name=CelebA,
description=CelebA}

\newglossaryentry{face_input}{%
name=\ensuremath{\boldsymbol{x}_I},
description={facial input image}
}
\newglossaryentry{context_input}{%
name=\ensuremath{\boldsymbol{x}_C},
description={context input image}
}
\newglossaryentry{model_input}{%
name=\ensuremath{\boldsymbol{x}},
description={input to the model}
}
\newglossaryentry{latent_space_dim}{%
name=\ensuremath{d},
description={latent space dimension}
}
\newglossaryentry{reconst}{%
name=\ensuremath{\hat{\boldsymbol{x}}},
description={reconstructed input}
}
\newglossaryentry{img_reconst}{%
name=\ensuremath{\hat{\boldsymbol{x}}_I},
description={reconstructed facial image}
}
\newglossaryentry{ctx_reconst}{%
name=\ensuremath{\hat{\boldsymbol{x}}_C},
description={reconstructed context}
}
\newglossaryentry{mean}{%
name=\ensuremath{\boldsymbol{\mu}},
description={mean of a Gaussian distribution}
}
\newglossaryentry{variance}{%
name=\ensuremath{\boldsymbol{\sigma}},
description={variance of a Gaussian distribution}
}
\newglossaryentry{img_mean}{%
name=\ensuremath{\boldsymbol{\mu}_I},
description={mean of a Gaussian distribution}
}
\newglossaryentry{img_variance}{%
name=\ensuremath{\boldsymbol{\sigma}_I},
description={variance of a Gaussian distribution}
}
\newglossaryentry{ctx_mean}{%
name=\ensuremath{\boldsymbol{\mu}_C},
description={mean of a Gaussian distribution}
}
\newglossaryentry{ctx_variance}{%
name=\ensuremath{\boldsymbol{\sigma}_C},
description={variance of a Gaussian distribution}
}
\newglossaryentry{shifted_mean}{%
name=\ensuremath{\boldsymbol{\mu}_S},
description={mean of a Gaussian distribution}
}
\newglossaryentry{shifted_variance}{%
name=\ensuremath{\boldsymbol{\sigma}_S},
description={variance of a Gaussian distribution}
}
\newglossaryentry{product_mean}{%
name=\ensuremath{\boldsymbol{\mu}_p},
description={mean of a Gaussian distribution}
}
\newglossaryentry{product_variance}{%
name=\ensuremath{\boldsymbol{\sigma}_p},
description={variance of a Gaussian distribution}
}
\newglossaryentry{latent_rep}{%
name=\ensuremath{\boldsymbol{z}},
description={latent features}
}
\newglossaryentry{latent_rep_shifted}{%
name=\ensuremath{\boldsymbol{z}_s},
description={latent features}
}
\newglossaryentry{y_true}{%
name=\ensuremath{y},
description={true class}
}
\newglossaryentry{y_pred}{%
name=\ensuremath{\hat{y}},
description={predicted class}
}
\newglossaryentry{cross_entropy}{%
name=\ensuremath{\text{CE}},
description={cross entropy}
}
\newglossaryentry{em_algo}{%
name=EM algorithm,
description={expectation-maximization algorithm, iterative method to obtain the maximum likelihood or maximum a posteriori estimate}
}

\newglossaryentry{model_params}{%
name=\ensuremath{\boldsymbol{\theta}},
description={model parameters}
}
\newglossaryentry{prob_input_model_params}{%
name=\ensuremath{P_{\boldsymbol{\theta}}(\boldsymbol{x})},
description={probability of input $x$ under model parameters $\theta$}
}
\newglossaryentry{prior_latent_model_params}{%
name=\ensuremath{P_{\boldsymbol{\theta}}(\boldsymbol{z})},
description={prior probability of latent representation $z$ under model parameters $\theta$}
}
\newglossaryentry{small_prior_latent_model_params}{%
name=\ensuremath{p_{\boldsymbol{\theta}}(\boldsymbol{z})},
description={prior probability of a single latent representation $z$ under model parameters $\theta$}
}
\newglossaryentry{prob_latent_input_model_params}{%
name=\ensuremath{P_{\boldsymbol{\theta}}(\boldsymbol{z}|\boldsymbol{x})},
description={probability of latent representation $z$ under model parameters $\theta$ and input $x$}
}
\newglossaryentry{prob_individual_input_model_params}{%
name=\ensuremath{P_{\boldsymbol{\theta}}(\boldsymbol{x}^1, \dotsc, \boldsymbol{x}_N)},
description={probability of input $x$ under model parameters $\theta$}
}
\newglossaryentry{joint_prob_input_i_latent_rep_model_params}{%
name=\ensuremath{p_{\boldsymbol{\theta}}(\boldsymbol{x}^i,\boldsymbol{z})},
description={probability of the $i$-th entry of input $x$ under model parameters $\theta$}
}
\newglossaryentry{prob_input_i_model_params}{%
name=\ensuremath{P_{\boldsymbol{\theta}}(\boldsymbol{x}^i)},
description={probability of the $i$-th entry of input $x$ under model parameters $\theta$}
}
\newglossaryentry{lower_prob_input_i_latent_rep_model_params}{%
name=\ensuremath{p_{\boldsymbol{\theta}}(\boldsymbol{x}^i|\boldsymbol{z})},
description={probability of the $i$-th entry of input $x$ under model parameters $\theta$}
}
\newglossaryentry{dataset_size}{%
name=\ensuremath{N},
description={size of the dataset}
}
\newglossaryentry{model_input_i}{%
name=\ensuremath{\boldsymbol{x}^i},
description={$i$-th entry of model input $x$}
}
\newglossaryentry{reconst_model_input_i}{%
name=\ensuremath{\boldsymbol{x}_{rec}^i},
description={$i$-th entry of model input $x$ reconstructed by the VAE decoder}
}
\newglossaryentry{latent_rep_i}{%
name=\ensuremath{\boldsymbol{z}^i},
description={latent representation of the $i$-th entry of input $x$}
}
\newglossaryentry{latent_rep_image}{%
name=\ensuremath{\boldsymbol{z}_I},
description={latent representation of the expression image}
}
\newglossaryentry{mean_i}{%
name=\ensuremath{\boldsymbol{\mu}^i},
description={mean of latent representation of the $i$-th entry of input $x$}
}
\newglossaryentry{variance_i}{%
name=\ensuremath{\boldsymbol{\sigma}^i},
description={variance of latent representation of the $i$-th entry of input $x$}
}
\newglossaryentry{assumed_posterior_latent_under_model_input_i}{%
name=\ensuremath{q_{\boldsymbol{\phi}}(\boldsymbol{z}|\boldsymbol{x}^i)},
description={assumed posterior of the latent representation $z$ under $i$-th entry of input $x$ and model parameters $\phi$}
}
\newglossaryentry{true_posterior_latent_under_model_input_i}{%
name=\ensuremath{p_{\boldsymbol{\phi}}(\boldsymbol{z}|\boldsymbol{x}^i)},
description={true posterior of the latent representation $z$ under $i$-th entry of input $x$ and model parameters $\phi$}
}
\newglossaryentry{likelihood_model_params_input_i}{%
name=\ensuremath{\mathcal{L}(\boldsymbol{\theta}, \phi; \boldsymbol{x}^i)},
description={likelihood of distribution parameters $\theta$ and $\phi$ under $i$-th entry of input $x$}
}
\newglossaryentry{posterior_params}{%
name=\ensuremath{\phi},
description={parameters of the true posterior}
}

\newglossaryentry{dec}{%
name=\ensuremath{Dec},
description={decoder}
}
\newglossaryentry{dec_i}{%
name=\ensuremath{Dec_I},
description={expression decoder decoder}
}
\newglossaryentry{dec_c}{%
name=\ensuremath{Dec_C},
description={context decoder}
}
\newglossaryentry{gen}{%
name=\ensuremath{Gen},
description={the full GAN generator, i.e. encoder and decoder applied to the input}
}
\newglossaryentry{dis}{%
name=\ensuremath{Dis},
description={discriminator}
}
\newglossaryentry{dis_i}{%
name=\ensuremath{Dis_I},
description={discriminator for the expression image}
}
\newglossaryentry{dis_c}{%
name=\ensuremath{Dis_C},
description={discriminator for the context}
}
\newglossaryentry{enc}{%
name=\ensuremath{Enc},
description={encoder}
}
\newglossaryentry{enc_i}{%
name=\ensuremath{Enc_I},
description={encoder of the facial image}
}
\newglossaryentry{enc_c}{%
name=\ensuremath{Enc_C},
description={encoder of the context}
}
\newglossaryentry{disc_loss}{%
name=\ensuremath{\mathcal{L}_{reconst}},
description={encoder}
}
\newglossaryentry{gen_loss}{%
name=\ensuremath{\mathcal{L}_{Gen}},
description={loss of the GAN generator}
}
\newglossaryentry{gan_loss}{%
name=\ensuremath{\mathcal{L}_{GAN}},
description={fully GAN loss}
}
\newglossaryentry{d_kl}{%
name=\ensuremath{D_{KL}},
description={KL-Divergence}
}
\newglossaryentry{prior_latent}{%
name=\ensuremath{p(\boldsymbol{z})},
description={prior over the latent variables}
}
\newglossaryentry{prior_latent_image}{%
name=\ensuremath{p(\boldsymbol{z}_I)},
description={prior over the image latent variables}
}
\newglossaryentry{posterior_latent_rep_image}{%
name=\ensuremath{p(\boldsymbol{z}_I|\boldsymbol{x}_I)},
description={probability of image latent representation $\boldsymbol{z}_I$ under input image $\boldsymbol{x}_I$}
}
\newglossaryentry{sampled_noise}{%
name=\ensuremath{\boldsymbol{z}_p},
description={noise sampled from a Gaussian distribution}
}
\newglossaryentry{posterior_class_under_latent}{%
name=\ensuremath{p(\boldsymbol{y}|\boldsymbol{z})},
description={probability of target class $y$ under latent representation $z$}
}

\newglossaryentry{prior_loss}{%
name=\ensuremath{\mathcal{L}_{prior}},
description={loss over the prior}
}
\newglossaryentry{dec_loss}{%
name=\ensuremath{\mathcal{L}_{Dec}},
description={loss for the VAE decoder}
}
\newglossaryentry{dis_loss}{%
name=\ensuremath{\mathcal{L}_{Dis}},
description={loss for the GAN discriminator}
}
\newglossaryentry{vae_gan_loss}{%
name=\ensuremath{\mathcal{L}_{joint}},
description={loss for the whole VAE-GAN architecture}
}
\newglossaryentry{prob_input}{%
name=\ensuremath{p(\boldsymbol{x})},
description={probability of the input $x$}
}
\newglossaryentry{posterior_input}{%
name=\ensuremath{q(\boldsymbol{x})},
description={probability of the input $x$}
}
\newglossaryentry{posterior_latent_rep_shifted_face_context}{%
name=\ensuremath{q(\boldsymbol{z}_s|\boldsymbol{x}_I,\boldsymbol{x}_C)},
description={posterior probability of latent features $z$ under input $x$}
}
\newglossaryentry{posterior_latent_input_image}{%
name=\ensuremath{q(\boldsymbol{z}_I|\boldsymbol{x}_I)},
description={posterior probability of latent image features $\boldsymbol{z}_I$ under image input $x_I$}
}
\newglossaryentry{hyper_params}{%
name=\ensuremath{hyper parameters},
description={Parameters needed to train a netural network, that are not network weights. Two exmaples are the batch size and learning rate. Both have to be chose in a sensible way but do not affect the network architecture, directly.}
}

\newglossaryentry{attention_map}{%
name=\ensuremath{\boldsymbol{A}},
description={attention map}
}
\newglossaryentry{attention_weights_mu_q}{%
name=\ensuremath{\boldsymbol{W}^{\mu}_q},
description={weights of layer computing the query for $\boldsymbol{\mu}$}
}
\newglossaryentry{attention_weights_mu_k}{%
name=\ensuremath{\boldsymbol{W}^{\mu}_k},
description={weights of layer computing the key for $\boldsymbol{\mu}$}
}
\newglossaryentry{attention_weights_mu_v}{%
name=\ensuremath{\boldsymbol{W}^{\mu}_v},
description={weights of layer computing the value for $\boldsymbol{\mu}$}
}
\newglossaryentry{query_mu}{%
name=\ensuremath{\boldsymbol{q}_{\mu}},
description={query of the attention mechanism}
}
\newglossaryentry{key_mu}{%
name=\ensuremath{\boldsymbol{k}_{\mu}},
description={key of the attention mechanism}
}
\newglossaryentry{value_mu}{%
name=\ensuremath{\boldsymbol{v}_{\mu}},
description={value of the attention mechanism}
}
\newglossaryentry{offset}{%
name=\ensuremath{\boldsymbol{O}},
description={offset computed by the context-attention network}
}
\newglossaryentry{mean_offset}{%
name=\ensuremath{\boldsymbol{o}_{\mu}},
description={offset for the mean}
}
\newglossaryentry{variance_offset}{%
name=\ensuremath{\boldsymbol{o}_{\sigma}},
description={offset for the variance}
}
\newglossaryentry{context_weight}{%
name=\ensuremath{m},
description={weight of the context offset during inference}
}
\newglossaryentry{weights_context_att_net}{%
name=\ensuremath{\boldsymbol{\theta}_{CAN}},
description={trainable weights of the context attention network}
}
\newglossaryentry{weights_classification_head}{%
name=\ensuremath{\boldsymbol{\theta}_{CH}},
description={trainable weights of the classification head}
}
\newglossaryentry{classification_head}{%
name=\ensuremath{CH},
description={classification head}
}
\newglossaryentry{softmax_class_probs}{%
name=\ensuremath{probs},
description={softmax class probabilities}
}

\newglossaryentry{mel_spec}{%
name=Mel spectrogram,
description={representation of sound data on a 2D plane which captures human hearing characteristics}
}
\newglossaryentry{target_leakage}{%
name=target leakage,
description={name of the effect when using data during training which is not available during testing, the model performs overly well and does not represent its performance on unseen data}
}
\newglossaryentry{iamge_net}{%
name=ImageNet,
description={standard dataset in computer vision to test a model's classification performance}
}
\newglossaryentry{tsne}{%
name=t-SNE,
description={method to analyze a set of vectors for their principle components so that they can be reduced to lower dimensions while retaining some characteristic information about the data}
}

\newif\ifreview
\reviewfalse

\ifreview
	\usepackage{lineno}

	\linenumbers
\fi

\begin{document}

%%%%%%%%%%%%%%%%%%%%% Add submission id, track, and title. %%%%%%%%%%%%%%%%%%%%%

% TODO: Please insert your submission number here
\def\SubNumber{9}

% TODO: Please uncomment the track this paper will be submitted to, comment all other lines
\def\GCPRTrack{Main Track}
%\def\GCPRTrack{Special Track: Pattern recognition in the life and natural sciences}
%\def\GCPRTrack{Special Track: Photogrammetry and remote sensing}
%\def\GCPRTrack{Young Researcher's Forum}
%\def\GCPRTrack{Fast Review Track}

% TODO: Replace with your title
\title{How Do You Perceive My Face?\\
Recognizing Facial Expressions in Multi-Modal Context by Modeling Mental Representations}

\titlerunning{How Do You Perceive My Face?}
% You can use \thanks for acknowledgment. Do not add any acknowledgment to the draft 
% version that is used for the review process.  
%\title{Title\thanks{XXX}}

\newcommand*\samethanks[1][\value{footnote}]{\footnotemark[#1]}

\ifreview
	% ANONYMOUS SUBMISSION FOR REVIEW
	% DO NOT MODIFY these for the draft version that is used for the review process.
	\titlerunning{GCPR 2024 Submission \SubNumber{}. CONFIDENTIAL REVIEW COPY.}
	\authorrunning{GCPR 2024 Submission \SubNumber{}. CONFIDENTIAL REVIEW COPY.}
	\author{GCPR 2024 - \GCPRTrack{}}
	\institute{Paper ID \SubNumber}
\else

	\author{Florian Blume\thanks{equal contribution}\inst{1,4}\orcidID{0000-0002-7557-1508}\faIcon{envelope} \and
	Runfeng Qu\samethanks\inst{1,4}\orcidID{0009-0008-7885-8812} \and
	Pia Bideau\inst{2}\orcidID{0000-0001-8145-1732} \and
    Martin Maier\inst{3,4}\orcidID{0000-0003-4564-9834} \and
    %%%%%% NOTE: "Abdel Rahman" is the last name, "Rasha" the given name
    Rasha Abdel Rahman\inst{3,4}\orcidID{0000-0002-8438-1570} \and
    Olaf Hellwich\inst{1,4}\orcidID{0000-0002-2871-9266}}
	
	\authorrunning{F. Blume et al.}
	
	\institute{
        Technische Universität Berlin
        \and Univ. Grenoble Alpes, Inria, CNRS, Grenoble INP, LJK
        \and Humboldt-Universität zu Berlin
        \and Science of Intelligence, Research Cluster of Excellence, Berlin
        \email{
            \\ \faIcon{envelope} florian.blume@tu-berlin.de, runfeng.qu@tu-berlin.de
        }
    }
\fi

\maketitle              % typeset the header of the contribution

\begin{abstract}
Facial expression perception in humans inherently relies on prior knowledge and contextual cues, contributing to efficient and flexible processing. For instance, multi-modal emotional context (such as voice color, affective text, body pose, etc.) can prompt people to perceive emotional expressions in objectively neutral faces. Drawing inspiration from this, we introduce a novel approach for facial expression classification that goes beyond simple classification tasks. Our model accurately classifies a perceived face and synthesizes the corresponding mental representation perceived by a human when observing a face in context. With this, our model offers visual insights into its internal decision-making process. We achieve this by learning two independent representations of content and context using a VAE-GAN architecture. Subsequently, we propose a novel attention mechanism for context-dependent feature adaptation. The adapted representation is used for classification and to generate a context-augmented expression. We evaluate synthesized expressions in a human study, showing that our model effectively produces approximations of human mental representations. We achieve State-of-the-Art classification accuracies of \oursmajority\% on the RAVDESS dataset and \oursmead\% on the MEAD dataset. We make our code publicly available\footnote{https://github.com/tub-cv-group/recognizing-by-modeling}.
\end{abstract}

\section{Introduction}
\label{sec:intro}

\begin{figure}[t]
    \centering
    \includegraphics[width=0.5\linewidth]{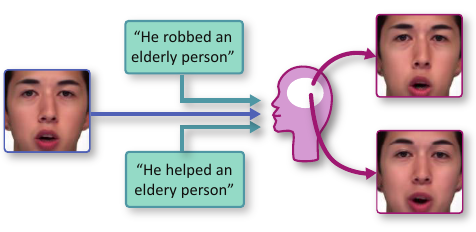}
    \caption{Visualization of the influence of context on \textbf{human perception} of facial expressions. The mental representation shifts congruently with context. % the valence. Several studies have investigated these dynamics \cite{russellReadingEmotionsFaces1997,suessPerceivingEmotionsNeutral2015,baumClearJudgmentsBased2020}. Generations of our model from \gls{ravdess} \cite{livingstoneRyersonAudioVisualDatabase2018} frames.
    }\label{sec:introduction:fig:overview:human_perception}
\end{figure}

Integrating multi-modal contextual information is crucial for generating adaptive behavior and enabling an agent to respond appropriately to its environment. More specifically, contextual information encompasses the multi-modal information that enhances the agent's perception and thus is a prerequisite for adaptive behavior. Latest work in cognitive psychology has shown that the human brain leverages contextual cues and prior knowledge to \textit{dynamically} adjust future predictions about incoming sensory input \cite{raoPredictiveCodingVisual1999,clarkWhateverNextPredictive2013}. Concurrent sound, textual cues, or prior knowledge offer additional information that shapes social perception~\cite{wieserFacesContextReview2012, maierKnowledgeaugmentedFacePerception2022}. For example, in the interpretation of facial expressions, the individual's voice plays a significant role in understanding their overall expression. As illustrated in \cref{sec:introduction:fig:overview:human_perception}, the same neutral face is perceived as displaying a more positive or negative expression when presented in the context of the respective prior beliefs about the person. In this work, we refer to the perceived facial expression as the \textit{synthesized mental representation}.

Previous works in computer vision typically either only perform \gls{fer} \cite{leeContextAwareEmotionRecognition2019,kostiContextBasedEmotion2019,mittalEmotiConContextAwareMultimodal2020,franceschiniMultimodalEmotionRecognition2022, mittalM3ERMultiplicativeMultimodal2020,chudasamaM2FNetMultimodalFusion2022} or generate expressions \cite{larsenAutoencodingPixelsUsing2016,bouzidFacialExpressionVideo2022,xuHighresolutionFaceSwapping2022,eskimezSpeechDrivenTalking2022,zhangJointExpressionSynthesis2022}. 
Only few approaches exist that perform both tasks jointly \cite{yanJointDeepLearning2020,sunUnsupervisedCrossViewFacial2023,zhangJointExpressionSynthesis2022,sadokMultimodalDynamicalVariational2024}. % but without a focus on generations as approximations of human mental representations.
Addressing recognition and generation jointly, however, poses an essential element in modelling human social interaction and creating effective communication between humans and artificial agents, enabling agents to mimic the expressions of conversational partners in a %fine-grained and 
context-sensitive manner and in direct alignment with the recognized expression. 

We propose a novel mechanism of fusing expressions and multi-modal context information encoded in the latent space of a \gls{vae} \cite{larsenAutoencodingPixelsUsing2016}. In particular, using an attention mechanism, our model dynamically adapts previously learned representations of facial expressions using context. Operating on lower-dimensional representations of facial expressions enables us to simultaneously produce well-aligned classifications and an approximation of the perceived expression. We verify the validity of these approximations in a rating study with 160 human observers and show SOTA classification accuracy on the \gls{ravdess} and \gls{mead} datasets. Our model performs a task that is similar to the studies in \cite{suessPerceivingEmotionsNeutral2015, baumClearJudgmentsBased2020,mullerIncongruenceEffectsCrossmodal2011}, where human participants were presented with individual photographs paired with affective-semantic contexts. Our contributions are threefold: \textbf{(1)} We present a model that for the first time simultaneously classifies expressions and produces approximations of their mental representations, which are inherently well aligned with the predicted class. \textbf{(2)} We evaluate these approximations in a human study. They capture the fine-grained effects of emotional context on human perception. \textbf{(3)} Our model is explainable due to its ability to visualize the adapted features by generating a context-augmented expression.

% Our approach is partly inspired by the ``reverse correlation'' technique in psychophysics \cite{
% manginiMakingIneffableExplicit2004,dotschReverseCorrelatingSocial2012}. In the context of face perception, reverse correlation typically involves presenting observers with a series of face images overlaid with random noise and asking them to make judgments about a particular facial characteristic (e.g., emotion, identity). However, the resulting classification images, considered approximations of internal mental representations serving as perceptual priors \cite{brinkmanVisualisingMentalRepresentations2017}, are subject to limitations. For instance, they remain noisy and require extensive human observations, which restricts the number of classification images typically generated in a study. To overcome these limitations, AI-supported techniques adopting a similar goal, but capable of producing high-quality images and handling a vast array of images and contexts, hold great potential for advancing research on social perception as well as human-machine interaction \cite{petersonDeepModelsSuperficial2022}. To achieve natural and effective interactions, artificial agents require not only advanced visual processing but also a closer alignment with the inherent human tendency to draw upon contextual cues \cite{fristonDuetOne2015,kirtayModelingRobotCorepresentation2020}.
\section{Related Works}
\label{sec:related_works}

We discuss three types of works: (1) context-sensitive \gls{fer}-only, (2) synthesizing expressions, and (3) performing both tasks jointly.
\\
\textbf{Multi-modal context-sensitive facial expression recognition.} Multi-modal context-sensitivity in \gls{fer} ranges from incorporating visual surroundings as additional information cues \cite{leeContextAwareEmotionRecognition2019,kostiEmotionRecognitionContext2017,kostiContextBasedEmotion2019,mittalEmotiConContextAwareMultimodal2020}, to drawing on audio \cite{chumachenkoSelfattentionFusionAudiovisual2022,zhangWeaklySupervisedVideo2023,fuCrossmodalFusionNetwork2021,dahmouniBimodalEmotionalRecognition2023,mocanuMultimodalEmotionRecognition2023,zhengTwoBirdsOne2023,luna-jimenezProposalMultimodalEmotion2021}, text \cite{zhangLearningEmotionRepresentations2023,zhengTwoBirdsOne2023}, body pose \cite{mittalEmotiConContextAwareMultimodal2020} or combining multiple context sources \cite{franceschiniMultimodalEmotionRecognition2022,chudasamaM2FNetMultimodalFusion2022,liDecoupledMultimodalDistilling2023}. Transformers have successfully been incorporated into \gls{fer} in multiple approaches \cite{chudasamaM2FNetMultimodalFusion2022,chumachenkoSelfattentionFusionAudiovisual2022,liDecoupledMultimodalDistilling2023,luna-jimenezProposalMultimodalEmotion2021}. Contrastive learning schemes have shown to extract general features that ensure good classification performance on unseen data \cite{chudasamaM2FNetMultimodalFusion2022,franceschiniMultimodalEmotionRecognition2022,zhangLearningEmotionRepresentations2023}. \citeauthor{franceschiniMultimodalEmotionRecognition2022} \cite{franceschiniMultimodalEmotionRecognition2022} outperform \gls{sota} methods on \gls{ravdess} using an unsupervised contrastive learning scheme on four modalities.  \citeauthor{luna-jimenezProposalMultimodalEmotion2021} \cite{luna-jimenezProposalMultimodalEmotion2021} use the transformer architecture and action units to predict expressions on \gls{ravdess}. In contrast to our work, generating context-augmented versions of the input face is not straightforward in these approaches.
\\
\textbf{Generating facial expressions in context.} \Glspl{gan} have been used for generating realistic looking face images \cite{larsenAutoencodingPixelsUsing2016,bouzidFacialExpressionVideo2022,xuHighresolutionFaceSwapping2022}. \citeauthor{larsenAutoencodingPixelsUsing2016} \cite{larsenAutoencodingPixelsUsing2016} combined them with a \gls{vae} to allow smooth transitions between representations. \citeauthor{fangFacialExpressionGAN2022} \cite{fangFacialExpressionGAN2022} employ a \gls{gan} to generate a talking face from audio. \citeauthor{pengEmoTalkSpeechDrivenEmotional2023} \cite{pengEmoTalkSpeechDrivenEmotional2023} generate a 3D talking head based on the audio of the \gls{ravdess} dataset, which is capable of producing facial expressions. Using artificial characters in a rating study results in a different perceptual experience for humans \cite{eiserbeckDeepfakeSmilesMatter2023}, which makes them inapplicable to our goal. \citeauthor{xuHighFidelityGeneralizedEmotional2023} \cite{xuHighFidelityGeneralizedEmotional2023} train their network in a CLIP-like \cite{radfordLearningTransferableVisual2021} fashion to generate sequences of talking faces. \citeauthor{Stypulkowski_2024_WACV} \cite{Stypulkowski_2024_WACV} employ latent diffusion \cite{rombachHighResolutionImageSynthesis2022} for the same task. None of these works target joint generation of mental representations and classification.
\\
\textbf{Joint Facial Expression Generation and Recognition.} Few works exist that perform the task of simultaneously generating facial expressions while also predicting expression classes. \citeauthor{sunUnsupervisedCrossViewFacial2023} \cite{sunUnsupervisedCrossViewFacial2023} train two \glspl{gan} cooperatively to recognize facial expressions under large view angle changes. \citeauthor{yanJointDeepLearning2020} \cite{yanJointDeepLearning2020} employ a \gls{gan} \cite{goodfellowGenerativeAdversarialNetworks2014} to overcome the lack of labeled training data in \gls{fer} by jointly training it together with an expression recognition network. Context sensitivity is not part of their work. \citeauthor{zhangJointExpressionSynthesis2022} \cite{zhangJointExpressionSynthesis2022} also draw on a \gls{gan}-based architecture and argue that by generating expressions, they help solve the issue of appearance variance in \gls{fer} and lack of training data. Their network processes only input images and does not take additional modalities into account. None of the disucssed works model mental representation through synthesized expressions.
\section{Method}

\begin{figure}[t]
    \centering
    \includegraphics[width=\textwidth]{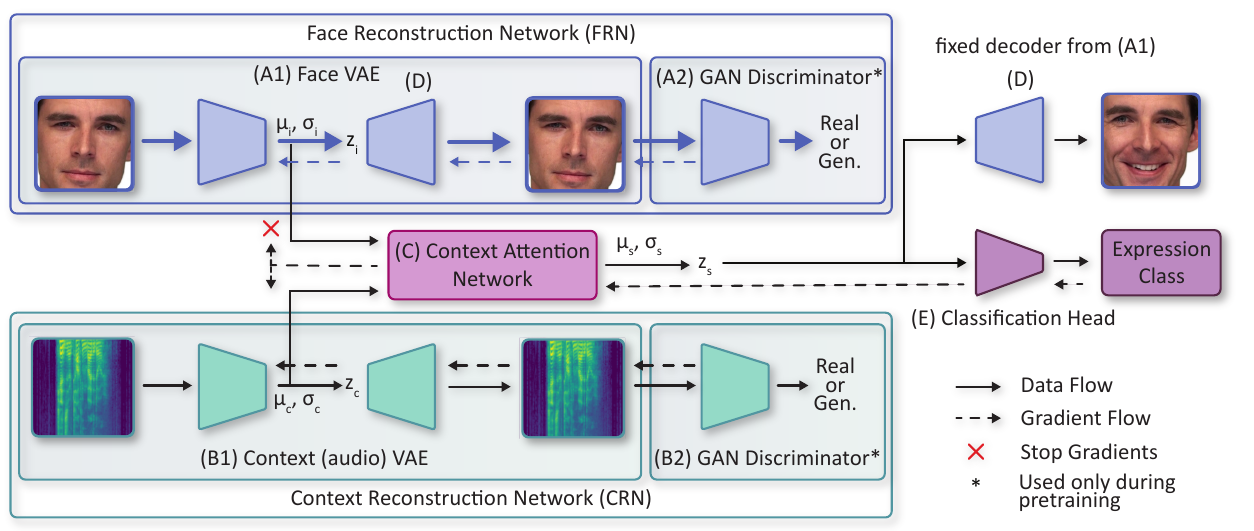}
    \caption{Overview of our full network architecture. The face and context reconstruction networks \acrshort{face_reconst_net} and \acrshort{context_reconst_net} are a \gls{vae}-\gls{gan} combination. The mean and variance of a facial input image and audio context (Mel spectrogram) are adapted by the \acrshort{context_att_net}, which shifts the face representation using the context representation. The classification head (E) classifies the shifted features and we use the fixed decoder of (A1) to visualize the expression. %Facial images taken from the \gls{ravdess} database \cite{livingstoneRyersonAudioVisualDatabase2018}.
    }
    \label{sec:method:pipeline_overview}
\end{figure}

In this Section, we describe our multi-modal approach that adapts an expression image using affective audio. Our proposed adaption module allows us to \textit{classify} a facial expression in light of context and at the same time \textit{synthesize} a novel facial expression as it would have been perceived by a human. We employ a two-stream encoder-decoder backbone, based on a \gls{vae}-\gls{gan} combination \cite{larsenAutoencodingPixelsUsing2016}, and an attention mechanism to combine their latent spaces in a context-sensitive way. \cref{sec:method:pipeline_overview} depicts an overview of this design: The representations learned by the \gls{face_reconst_net} and the \gls{context_reconst_net} are adapted in the \gls{context_att_net}, by shifting the facial features using the context features. The shifted representation is visualized using the fixed decoder (D) of the \gls{face_reconst_net} and classified with the classification head (E). Our model operates and is trained on individual frames together with audio context. To evaluate our model on videos, we perform majority voting over classes of the frames.

\subsection{Face and Context Reconstruction Network}

The face and context reconstruction networks (called \gls{face_reconst_net} and \gls{context_reconst_net} respectively) both consist of a \gls{vae} and \gls{gan} discriminator. We follow \cite{larsenAutoencodingPixelsUsing2016} and add a \gls{gan} discriminator for training to increase image quality.
\\
\textbf{VAE Module.} Similarly to \cite{huangIntroVAEIntrospectiveVariational2018}, we add skip connections to the en- and decoder of the \glspl{vae} to speed up the training process and allow processing of larger image resolutions. Let \gls{enc_i} be the expression encoder of (A1) and $\gls{face_input} \in \mathbb{R}^{m \times n \times 3}$ an input expression image:
\begin{equation}
    (\gls{img_mean}, \gls{img_variance}) = \gls{enc_i}(\gls{face_input})
\end{equation}
where $\gls{img_mean} \in \mathbb{R}^{\gls{latent_space_dim}}$ and $\gls{img_variance} \in \mathbb{R}^{\gls{latent_space_dim}}$ denote the mean and variance of a Gaussian distribution, respectively. Mean $\gls{ctx_mean} \in \mathbb{R}^{\gls{latent_space_dim}}$ and $\gls{ctx_variance} \in \mathbb{R}^{\gls{latent_space_dim}}$ variance of the context Mel spectrogram $\gls{context_input} \in \mathbb{R}^{u \times v}$ are computed using the encoder of (B1) and the following formulas are applied analogously.

The \textbf{prior loss} term keeps the latent distribution close to a Gaussian:
\begin{equation} \label{ch:approximating_representations:sec:pregc:sec:methods:eqn:prior_loss}
    \gls{prior_loss} = \gls{d_kl}( \gls{posterior_latent_input_image} || \gls{prior_latent_image})
\end{equation}
\gls{d_kl} is the \gls{kl_divergence}, \gls{posterior_latent_input_image} is the posterior of the latent vector $\gls{latent_rep_image} \in \mathbb{R}^{\gls{latent_space_dim}}$ under input \gls{face_input} and \gls{prior_latent_image} is the Gaussian prior over the latent vector.

The \textbf{reconstruction loss} term penalizes the feature map of the discriminator (A2 and B2 in \cref{sec:method:pipeline_overview}) at a certain level, as proposed in \cite{larsenAutoencodingPixelsUsing2016}, using \gls{mse}:

\begin{equation} \label{ch:approximating_representations:sec:pregc:sec:methods:eqn:target_loss}
    \gls{disc_loss} = \gls{mse}(\gls{dis_i}^l(\gls{dec_i}(\gls{enc_i}(\gls{face_input}))), \gls{dis_i}^l(\gls{face_input}))
\end{equation}
where \gls{face_input} is the input image and $\gls{dis_i}^l$ the $l$-th feature map of the discriminator.
\\
\textbf{GAN Module.} The \gls{gan} discriminator's task is to distinguish between input images \gls{face_input} from the dataset and reconstructions \gls{img_reconst}. In addition, it is tasked to identify reconstructions from random noise $\gls{sampled_noise} \sim \mathcal{N}(0, 1)$ to enforce generation capabilities in the \gls{vae}. The overall loss is the following:

\begin{align}
    \gls{gan_loss} = & \log (\gls{dis_i}(\gls{face_input})) + \log (1 - \gls{dis_i}(\gls{dec_i}(\gls{enc_i}(\gls{face_input}))))  \notag \\
    & + \log (1 - \gls{dis_i}(\gls{dec_i}(\gls{sampled_noise})))
\end{align}

\paragraph{Joint Training.} In the pretraining phase, \gls{face_reconst_net} and \gls{context_reconst_net} are trained unsupervised for input reconstruction and are fixed in subsequent training. We follow the training algorithm of \cite{larsenAutoencodingPixelsUsing2016} and compute the joint update as follows:

\begin{align} 
    \boldsymbol{\theta}_{\gls{enc}} & \stackrel{+}\longleftarrow - \nabla_{\boldsymbol{\theta}_{\gls{enc}}} \left(\beta \gls{prior_loss} + \gls{disc_loss}\right) \label{ch:approximating_representations:sec:pregc:sec:methods:eqn:vae_encoder_loss} \\
    \boldsymbol{\theta}_{\gls{dec_i}} & \stackrel{+}\longleftarrow - \nabla_{\boldsymbol{\theta}_{\gls{dec_i}}} \left( \gls{disc_loss} - \gls{gan_loss} \right) \\
    \boldsymbol{\theta}_{\gls{dis_i}} & \stackrel{+}\longleftarrow - \nabla_{\boldsymbol{\theta}_{\gls{dis_i}}} \gls{gan_loss}
\end{align}
where $\beta \in \mathbb{R}$ is a hyperparameter to weigh the prior loss.

\subsection{Context-Attention Network (CAN) and Classification Head}
\begin{figure}[b]
    \centering
    \includegraphics[width=0.5\linewidth]{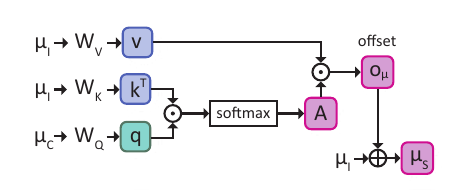}
    \caption{Detailed view of the \gls{context_att_net} from \cref{sec:method:pipeline_overview} for adapting the means. $\odot$ is element-wise multiplication, $\oplus$ addition. We left out index $\mu$ on the weights for simplicity. The shifted variance \gls{shifted_variance} is computed analogously.}\label{sec:methods:fig:context_attention_computation}
\end{figure}
Our proposed attention mechanism shifts the facial expression distribution of the \gls{face_reconst_net} based on the context distribution of the \gls{context_reconst_net}. \cref{sec:methods:fig:context_attention_computation} illustrates this fusion technique.
\\
\textbf{Context-Attention Net.} The \gls{context_att_net} computes attention maps based on mean and variance of the distributions of the context and the facial expression input. We use these maps to compute offsets $\gls{mean_offset} \in \mathbb{R}^{\gls{latent_space_dim}}$ and $\gls{variance_offset} \in \mathbb{R}^{\gls{latent_space_dim}}$ to shift the face mean and variance context-dependently. We compute the following parameters for the attention mechanism:
\begin{align}
    \gls{query_mu} = \gls{attention_weights_mu_q} \gls{ctx_mean} \\
    \gls{key_mu} = \gls{attention_weights_mu_k} \gls{img_mean} \\
    \gls{value_mu} = \gls{attention_weights_mu_v} \gls{img_mean}
\end{align}
$\gls{attention_weights_mu_q}, \gls{attention_weights_mu_k},\gls{attention_weights_mu_v} \in \mathbb{R}^{\gls{latent_space_dim} \times \gls{latent_space_dim}}$ are the trainable parameters of linear layers (bias omitted for simplicity). We compute the attention map $\gls{attention_map} \in \mathbb{R}^{\gls{latent_space_dim} \times \gls{latent_space_dim}}$ as

\begin{equation}
    \gls{attention_map} = \text{softmax}(\gls{query_mu} \gls{key_mu}^T)
\end{equation}
where the softmax function is applied row-wise. Note that we reverse the attention mechanism from \cite{NIPS2017_3f5ee243} - we do not compute the dot products of the query with all keys, instead, we compute the dot product of a key to all queries. We do this to get attention on the facial mean based on the context mean. The offset $\gls{mean_offset}$ and the resulting shifted mean \gls{shifted_mean} are then computed as 
\begin{align}
    \gls{mean_offset} & =  \gls{attention_map} \gls{value_mu} \\
    \gls{shifted_mean} & = \gls{mean_offset} + \gls{img_mean}
\end{align}
\cref{sec:methods:fig:context_attention_computation} provides a visualization of these relationships. During inference we can vary the strength of the context influence by multiplying the offset with a weight, to allow smooth modulation of the offset:
\begin{equation}
    \gls{shifted_mean} = \gls{context_weight} \cdot \gls{mean_offset}+ \gls{img_mean}\label{sec:methods:eqn:inference_context_weighting}
\end{equation}
We compute the new (shifted) variance \gls{shifted_variance} analogously, the only difference being that we operate in log scale.
\\
\textbf{Joint Context Attention Network and Classification Head Training.} To train the \gls{context_att_net} and classification head jointly, we first initialize the latter by training it directly on the facial features using the expression classes and cross-entropy loss. Next, we train the network together using the following loss:
\begin{equation}
    \mathcal{L} = \gls{cross_entropy}(\text{E}(\gls{shifted_mean}), \gls{y_true}) + \alpha \gls{d_kl}(\gls{posterior_latent_rep_image}||\gls{posterior_latent_rep_shifted_face_context})\label{ch:approximating_representations:sec:pregc:sec:methods:eqn:joint_class_loss}
\end{equation}

\noindent
where \gls{cross_entropy} is the cross-entropy, E is the classification head and $\alpha \in \mathbb{R}$ is a hyperparameter to regularize the shift. Furthermore, during training, we propose a novel data augmentation approach for multi-modal settings, where we swap contexts for a specific actor within its expression class.\footnote{Note, this data augmentation is only possible for multi-modal datasets, where content comes with different context variations within their respective expression class.}
\section{Experiments}
We evaluate our approach on publicly available datasets~\cite{livingstoneRyersonAudioVisualDatabase2018,wangMEADLargeScaleAudioVisual2020}. We discuss the results in Section~\ref{sec:results-classification} and \ref{sec:results-mentalrepresenations}. An ablation study is shown to support the understanding of our proposed approach for \gls{fer} in multi-modal context.

\subsection{Datasets}

\textbf{\gls{celeb_db} (Pretraining).} We pretrain the \gls{face_reconst_net} unsupervisedly for face reconstruction on the large-scale dataset \gls{celeb_db} \cite{liuDeepLearningFace2015}. \gls{celeb_db} is a prevalent dataset for face attribute recognition and consists of roughly 200k images showing 10k different identities. The \gls{context_reconst_net} is trained using the respective downstream datasets.
\\
\textbf{\gls{ravdess} (Downstream).} \gls{ravdess} \cite{livingstoneRyersonAudioVisualDatabase2018} consists of videos of 24 identities. Each video is labeled with one of the seven expression classes calm, happy, sad, angry, fearful, surprise, and disgust, with an additional binary label for the intensity. We extract 16 frames from each video at regular intervals and use the video's label for each. Following~\cite{chumachenkoSelfattentionFusionAudiovisual2022} the neutral class is omitted to reduce noise.
\\
\textbf{\gls{mead} (Downstream).} \gls{mead} \cite{wangMEADLargeScaleAudioVisual2020} is a large-scale dataset targeting talking-face generation, which also features \gls{fer} labels. Similar to \cite{livingstoneRyersonAudioVisualDatabase2018}, \gls{mead} contains videos of 60 actors speaking with different emotions at different intensity levels. We use the frontal view recordings as proposed by~\cite{sadokMultimodalDynamicalVariational2024} and apply the same frame extraction approach as for \gls{ravdess}.

\subsection{Implementation Details}\label{sec:experiments:sec:implementation_details}

We set the batch size to 256, $\beta$ (\cref{ch:approximating_representations:sec:pregc:sec:methods:eqn:vae_encoder_loss}) and $\alpha$ (\cref{ch:approximating_representations:sec:pregc:sec:methods:eqn:joint_class_loss}) to 0.00001 and $\gls{context_weight}=1.0$ (\cref{sec:methods:eqn:inference_context_weighting}). The dimension of the latent space of the \gls{face_reconst_net} and \gls{context_reconst_net} is $\gls{latent_space_dim} = 512$. We use MTCNN \cite{zhangJointFaceDetection2016} to detect faces. We then resize them to $128 \times 128$ pixels, which is also the size of the reconstructed and generated images. We apply random horizontal flipping as data augmentation. For generating the Mel spectrograms, we chose 128 Mel bins, a sample rate of 22050, a window and FFT length of 1310, and a hop length of 755. We use the Adam optimizer \cite{kingmaAdamMethodStochastic2017} with a learning rate of $0.00003$ and a weight decay of $0.01$.
\\
\textbf{Reconstruction Pretraining.} We pretrain the \gls{face_reconst_net} unsupervised for facial image reconstruction on \gls{celeb_db} and the \gls{context_reconst_net} for context reconstruction on the Mel spectrograms of the downstream datasets. Pretraining is run for 400 epochs. The learning rate is decreased by factor 10 after 150 and 300 epochs.
\\
\textbf{Downstream Classification Training.}  During downstream training, the \gls{face_reconst_net} is fixed and the last two layers of the \gls{context_reconst_net} are fine-tuned. We first initialize the one-layer classification head by training it directly on the facial features of the \gls{face_reconst_net} to obtain a suitable initialization for its weights. Next, we train the \gls{context_att_net} and the single-layer classification head jointly together using the loss from \cref{ch:approximating_representations:sec:pregc:sec:methods:eqn:joint_class_loss}. %To improve generalization, we paired input images of a person with a random audio file of the same person with matching class (\ie intensity, trial, and the spoken sentence are varied). During testing, we chose the actual respective audio file. 
Note that the decoder from (A1), which we use to visualize the shifted expression, is fixed and not trained in this step.

For both \gls{ravdess} and \gls{mead}, we performed \textit{k}-fold cross validation with $k=10$, similarly to other works \cite{franceschiniMultimodalEmotionRecognition2022,mocanuMultimodalEmotionRecognition2023}, splitting the folds along the identities (\ie one identity can only occur in test, validation or train set).

\subsection{Facial Expression Recognition Performance}
\label{sec:results-classification}

\begin{table}[t]
    \small
    \centering
    \caption{Classification accuracy of \gls{sota} methods on \textbf{\gls{ravdess}}. V = videos, A = audio, K = facial keypoints. Gen and Class are generative and classification approaches, respectively. Bold are best results in the respective group.}
    \label{sec:methods:sec:downstream_classification:tab:ravdess_accuracy}
    \begin{tabular}{lccccc}
        \hline
        Model & Modalities & Year & Gen & Class & Acc \\ \hline
        \multicolumn{6}{c}{\textit{Classification-Only Approaches}} \\ \hline
        \citeauthor{franceschiniMultimodalEmotionRecognition2022} \cite{franceschiniMultimodalEmotionRecognition2022} & A + V + K & 2022 &  & \cmark & 78.54 \\ \hline
        \citeauthor{fuCrossmodalFusionNetwork2021} \cite{fuCrossmodalFusionNetwork2021} & A + V & 2021 &  & \cmark & 75.76 \\
        \citeauthor{ghalebMultimodalAttentionMechanismTemporal2020} \cite{ghalebMultimodalAttentionMechanismTemporal2020} & A + V & 2020 & & \cmark & 76.30 \\
        \citeauthor{chumachenkoSelfattentionFusionAudiovisual2022} \cite{chumachenkoSelfattentionFusionAudiovisual2022} & A + V & 2022 &  & \cmark & 81.58 \\ 
        \citeauthor{dahmouniBimodalEmotionalRecognition2023} \cite{dahmouniBimodalEmotionalRecognition2023} & A + V & 2023 &  & \cmark & \textbf{85.76} \\ \hline
        \multicolumn{6}{c}{\textit{Joint Classification-Generation Approaches}} \\ \hline 
        \citeauthor{sadokMultimodalDynamicalVariational2024} \cite{sadokMultimodalDynamicalVariational2024} & A + V & 2024 &  \cmark & \cmark & 68.8 \\
        Ours & A + V & 2024 & \cmark & \cmark & \textbf{81.01}\\ \hline
    \end{tabular}
\end{table}

\begin{table}[t]
    \small
    \centering
    \caption{Classification accuracy of \gls{sota} methods on \textbf{\gls{mead}}.}
    \label{sec:methods:sec:downstream_classification:tab:mead_accuracy}
    \begin{tabular}{lccccc}
        \hline
        Model & Modalities & Year & Gen & Class & Acc \\ \hline
        wav2vec \cite{schneider19_interspeech} & A & 2019 &  & \cmark & 68.4 \\
        \citeauthor{sadokMultimodalDynamicalVariational2024} \cite{sadokMultimodalDynamicalVariational2024} & A + V & 2024 &  \cmark & \cmark & 73.2 \\
        Ours & A + V & 2024 & \cmark & \cmark & \textbf{79.0}\\ \hline
    \end{tabular}
\end{table}

We provide classification results for \gls{ravdess} in \cref{sec:methods:sec:downstream_classification:tab:ravdess_accuracy}, and for \gls{mead} in \cref{sec:methods:sec:downstream_classification:tab:mead_accuracy}. All compared methods use a dataset split by identities, ensuring no identity seen during training appears during testing.
%Both tables are limited to approaches that split the datasets along identities and not videos. 
For our method, the final prediction for a test video is obtained by majority voting across frames. %, \ie the most frequently occurring class wins. 
Our model achieves an accuracy of \oursmajority\% on \gls{ravdess} and \oursmead\% on \gls{mead}, matching SOTA performance on classical \gls{fer}. We largely outperform methods - \ie by 17.75\% on \gls{ravdess} and 8.47\% on \gls{mead}, that tackle the dual problem of classification and the synthesize of the corresponding percept.  
%Our work and \cite{sadokMultimodalDynamicalVariational2024} are the only two that jointly classify and generate expressions on \gls{ravdess} and \gls{mead}, while we outperform them.

%Note that as mentioned before, achieving the highest classification accuracy was not our primary goal. Our approach of a human-centered view on \gls{fer} is in line with recent opinions in machine learning demanding to diverge from purely leaderboard-driven research \cite{maierKnowledgeaugmentedFacePerception2022}.

The influence of context on the final \gls{fer} accuracy is highlighted by visualizing the per-class accuracy in confusion matrices for two conditions: in \cref{sec:experiments:sec:downstream_classification_task:fig:confusion_matrices:fig:face_face}, the \gls{context_att_net} computed the offset for the facial features based on the facial (instead of the context) features to simulate missing context. In \cref{sec:experiments:sec:downstream_classification_task:fig:confusion_matrices:fig:face_context}, it received the face together with the context features, as intended by our approach. The latter exhibits higher (or equal) probabilities on the diagonal for every class. This proves that the reported accuracy cannot be attributed solely to the computational capabilities of the \gls{context_att_net} but is a result of the meaningful adaption procedure.

\def\confmatsize{0.8}
\def\tsneconfmatsubfigsize{0.49}
\pgfplotsset{colormap/jet}

\begin{figure}[t]
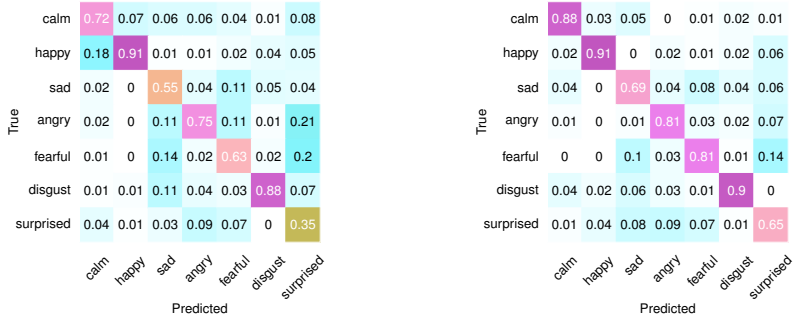

    \begin{subfigure}[t]{\tsneconfmatsubfigsize\linewidth}
        \centering
        \confmatrix{sec/data/confusion_matrix/confusion_matrix_face_face.csv}
        \caption{Confusion matrix of \gls{context_att_net}, face features shifted with face features.}\label{sec:experiments:sec:downstream_classification_task:fig:confusion_matrices:fig:face_face}
    \end{subfigure}
    \hfill
    \begin{subfigure}[t]{\tsneconfmatsubfigsize\linewidth}
        \centering
        \confmatrix{sec/data/confusion_matrix/confusion_matrix_face_context.csv}
        \caption{Confusion matrix of \gls{context_att_net}, face features shifted with context features.}\label{sec:experiments:sec:downstream_classification_task:fig:confusion_matrices:fig:face_context}
    \end{subfigure}
    \caption{Confusion matrices for our model. (a) Uni-modal setting with simulated missing context. (b) Multi-modal setting showing the clear diagonal.}\label{sec:experiments:sec:downstream_classification_task:fig:confusion_matrices}
\end{figure}

\begin{figure}[t]
    \vspace{-3.7cm}
    \hfill
    \centering
    \begin{subfigure}[t]{0.3\linewidth}
        \centering
        \begin{tikzpicture}[font=\tiny]
            \tikzset{mark options={mark size=3}}
            \begin{axis}[
                width=1.0\linewidth,
                height=1.0\linewidth,
                cycle list name=scatter,
                axis line style={draw=none},
                yticklabels={,,},
                xticklabels={,,},
                ticks=none,
                scatter,
                scatter src=explicit symbolic,
                scatter/classes={
                    01={mark=*},
                    02={mark=triangle*,mark size=4},
                    03={mark=square*}
                }
            ]
                \foreach \i in {
                    0,
                    1,
                    2,
                    3,
                    4,
                    5,
                    6%
                }{
                    \addplot+[
                        only marks,
                        x filter/.expression={\thisrow{class}==\i ? x : nan}
                    ] table[x=Dim1, y=Dim2, meta=id, col sep=comma] {sec/data/tsne/original_features_219.csv};
                }
            \end{axis}
        \end{tikzpicture}
        \caption{Face-only latent features obtained from the \gls{face_reconst_net} without taking context into account.} \label{sec:experiments:sec:downstream_classification:fig:tsne:fig:face_only}
    \end{subfigure}
    \hfill
    \begin{subfigure}[t]{0.3\linewidth}
        \centering
        \begin{tikzpicture}[font=\tiny]
            \tikzset{mark options={mark size=3}}
            \begin{axis}[
                width=1.0\linewidth,
                height=1.0\linewidth,
                cycle list name=scatter,
                axis line style={draw=none},
                yticklabels={,,},
                xticklabels={,,},
                ticks=none,
                scatter,
                scatter src=explicit symbolic,
                scatter/classes={
                    01={mark=*},
                    02={mark=triangle*,mark size=4},
                    03={mark=square*}
                }
            ]
                \foreach \i in {
                    0,
                    1,
                    2,
                    3,
                    4,
                    5,
                    6%
                }{
                    \addplot+[only marks,x filter/.expression={\thisrow{class}==\i ? x : nan}] table[x=Dim1, y=Dim2, meta=id, col sep=comma] {sec/data/tsne/shifted_features_face_only.csv};
                }
            \end{axis}
        \end{tikzpicture}
        \caption{Features obtained from the \gls{context_att_net} by providing the face features again as context.} \label{sec:experiments:sec:downstream_classification:fig:tsne:fig:face_face}
    \end{subfigure}
    \hfill
    \begin{subfigure}[t]{0.3\linewidth}
        \centering
        \begin{tikzpicture}[font=\tiny]
                \tikzset{mark options={mark size=3}}
                \begin{axis}[
                    width=1.0\linewidth,
                    height=1.0\linewidth,
                    cycle list name=scatter,
                    axis line style={draw=none},
                    yticklabels={,,},
                    xticklabels={,,},
                    ticks=none,
                    scatter,
                    scatter src=explicit symbolic,
                    scatter/classes={
                        01={mark=*},
                        02={mark=triangle*,mark size=4},
                        03={mark=square*}
                    },
                    legend entries={
                        01,
                        02,
                        03%
                    },
                    legend style = {
                        font=\tiny\sansmath\sffamily,
                        draw=none,
                        xshift=0pt,
                        mark options={solid,fill=black,draw=black}
                    },
                    legend pos=outer north east,
                ]
                    % \addplot+[only marks] table [
                    %     x expr=nan,
                    %     y expr=nan,
                    % ] {sec/data/tsne/shifted_features_220.csv};
                    \foreach \i in {
                        0,
                        1,
                        2,
                        3,
                        4,
                        5,
                        6%
                    }{
                        \addplot+[only marks,x filter/.expression={\thisrow{class}==\i ? x : nan}] table[x=Dim1, y=Dim2, meta=id, col sep=comma] {sec/data/tsne/shifted_features_219.csv};
                    }
                    \addplot+[only marks,x filter/.expression={\thisrow{class}==1 ? x : nan}] table[x=Dim1, y=Dim2, meta=id, col sep=comma] {sec/data/tsne/shifted_features_219.csv};
                    
                \end{axis}
                    % this is a dummy `axis' environment only to create the second legend
                \begin{axis}[
                    % set some axis limits and plot the coordinates outside that box
                    % so they don't show up
                    xmin=-25.34,
                    xmax=-24.34,
                    ymin=75.1349,
                    ymax=76.1349,
                    % of course we also don't want to show this axis
                    hide axis,
                    cycle list name=scatter,
                    % we need only marks
                    only marks,
                    % state the legend entries for the second legend
                    % (here we don't have scatter classes, so each `\addplot' gets its
                    %  own entry in the legend)
                    legend entries={
                        calm,
                        happy,
                        angry,
                        surprised,
                        disgust,
                        sad,
                        fearful%
                    },
                    % place it below the other legend
                    % therefore we have to shift it down (manually)
                    legend pos=outer north east,
                    legend style={
                        yshift=-102pt,
                        xshift=-118pt,
                        font=\tiny\sansmath\sffamily,
                        draw=none
                    },
                ]
                    % just add some dummy plots to create the legend
                    \foreach \i in {0,...,6} {
                        \addplot+ [mark=*] coordinates { (0,0) };
                    }
                \end{axis}
        \end{tikzpicture}
        \caption{\textbf{Proposed approach}: Features obtained from the \gls{context_att_net} by shifting using the context.} \label{sec:experiments:sec:downstream_classification:fig:tsne:fig:face_context}
    \end{subfigure}
    \hfill
    \caption{Comparison of \gls{tsne} visualizations of all samples (\ie frames plus audio) from the \gls{ravdess} identities 01, 02 and 03.}\label{sec:experiments:sec:downstream_classification:fig:tsne}
\end{figure}
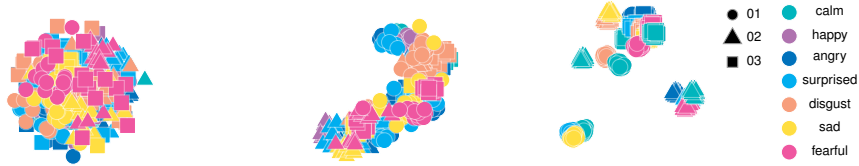

The \gls{tsne} plots in \cref{sec:experiments:sec:downstream_classification:fig:tsne} visualize the structure of the latent space in different conditions: (1) The face-only features as we receive them from the \gls{face_reconst_net} do not cluster in any particular way (\cref{sec:experiments:sec:downstream_classification:fig:tsne:fig:face_only}). (2) Employing our \gls{context_att_net} but providing the face features twice instead of combining with the context leads to a more structured latent space (\cref{sec:experiments:sec:downstream_classification:fig:tsne:fig:face_face}) and (3) taking the audio context into account leads to a clustered latent space that makes classification easier (\cref{sec:experiments:sec:downstream_classification:fig:tsne:fig:face_context}).

\begin{table}[t]
    \small
    \centering
    \caption{Performance ablation study on \gls{ravdess} of a VGG16 network, a single linear layer on the features of the \gls{face_reconst_net} and \gls{context_reconst_net}, and the \gls{context_att_net}.} 
    \begin{tabular}{lcc}
    \hline
        Model & Modality & Acc. \% \textdownarrow{} \\ \hline
        Baseline VGG16 \cite{simonyanVeryDeepConvolutional2015} & Video & 64.40 \\
        Single-Layer & Video & 68.06 \\
        \gls{context_att_net} & Video & 68.99 \\ \hline
        Single-Layer & Audio & 48.66 \\ 
        \gls{context_att_net} & Audio & 50.05 \\ \hline
        CAN (strict audio) & Video + Audio & 79.64 \\
        CAN (Ours) & Video + Audio & \textbf{\oursmajority} \\ \hline\end{tabular}\label{sec:methods:sec:downstream_classification:tab:accuracy_ablation}
\end{table}

\textbf{Ablation Study.} \cref{sec:methods:sec:downstream_classification:tab:accuracy_ablation} lists classification performance of our \gls{context_att_net} and its components. We assess the quality of face and context features independently by classifying learned features directly using a single-layer classifier. Experiments for the \gls{context_att_net} in a simulated uni-modal mode are run by providing face or context twice. If no context is provided, the \gls{context_att_net} performs similarly to the single layer classifier using learned representations. However, when learned features are adapted to context, the final classification performance improves by 15.44\% (strict audio). Additionally, our new data augmentation technique for multi-modal settings further enhances performance by 1.72\%.
%and evaluated the full model's performance for pairing frames with the actual audio, \ie not varying intensity, trial, etc.~(\textit{strict audio}).  All are lower than the \oursmajority\% achieved by our full model.

\subsection{Mental Representations: Context-Augmented Expressions}
\label{sec:results-mentalrepresenations}

\begin{figure}[t]
    \centering\includegraphics[width=0.63\linewidth]{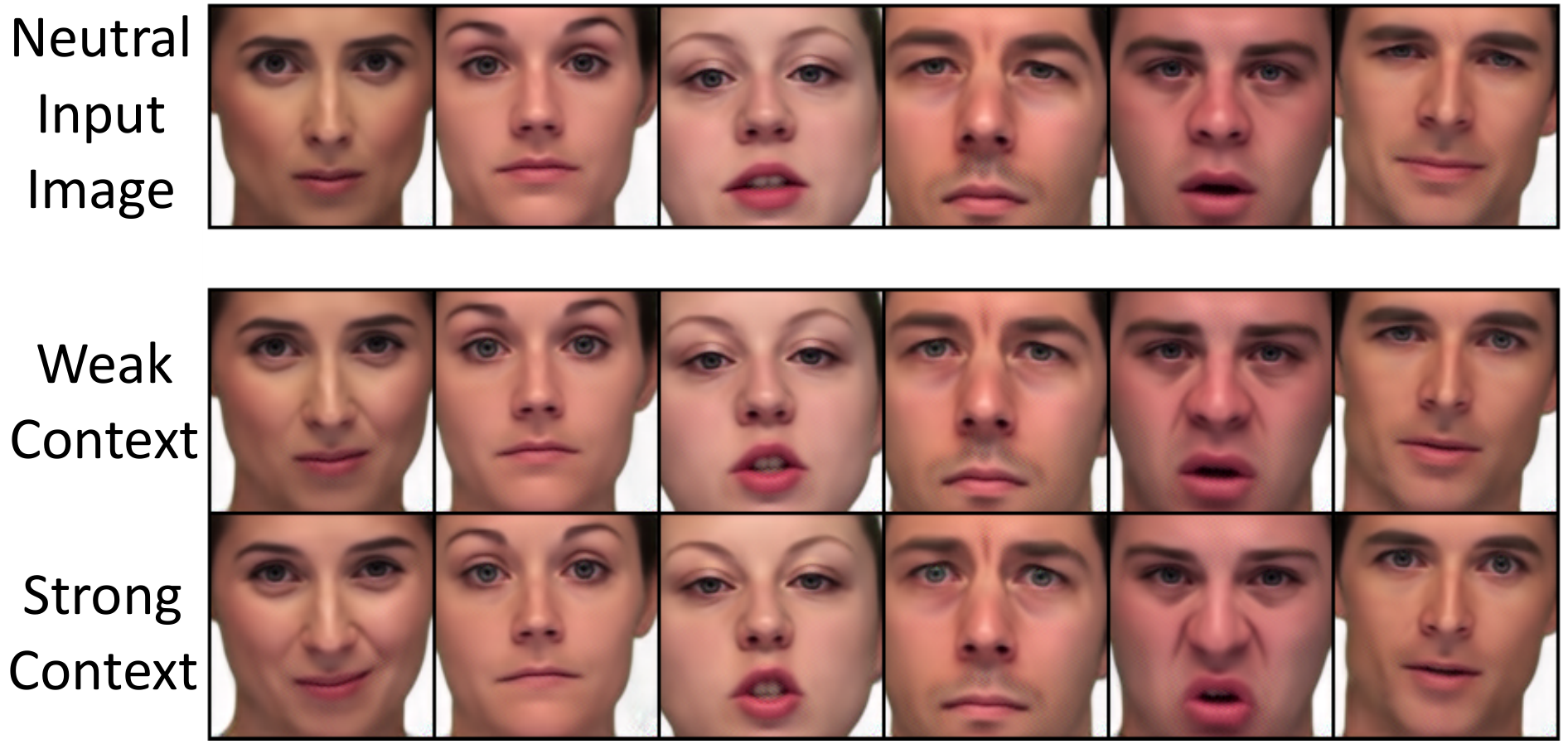}
    \caption{\textbf{Neutral} faces generated with intensity variations as provided by \gls{ravdess}. Left to right: happy, sad, angry, fearful, disgusted, surprised.} \label{sec:experiments:sec:context_augmented_generations:fig:context_weak_strong}
\end{figure}

We show qualitative results of our approach on \gls{ravdess} in different conditions (additional generations for \gls{ravdess} and \gls{mead} in the suppl. material). To visualize the adapted features obtained from the \gls{context_att_net}, we use the decoder (D) from \cref{sec:method:pipeline_overview}. \cref{sec:experiments:sec:context_augmented_generations:fig:context_weak_strong} depicts the effect of the two kinds of intensities provided in \gls{ravdess} on neural input faces. The stronger the intensity, the stronger the facial expression in the generated image. The strength of the effect is subtle but resembles human perception~\cite{maierKnowledgeaugmentedFacePerception2022}, which is our goal. % since they are also not extensively strong in human perception \cite{maierKnowledgeaugmentedFacePerception2022}.

To our knowledge, the only publications showing generations on \gls{ravdess} are \citeauthor{sinhaEmotionControllableGeneralizedTalking2022} \cite{sinhaEmotionControllableGeneralizedTalking2022}, \citeauthor{maDreamTalkWhenExpressive2023} \cite{maDreamTalkWhenExpressive2023} and \citeauthor{fangFacialExpressionGAN2022} \cite{fangFacialExpressionGAN2022}. They aim at rendering an input face that strongly resembles the emotion in the accompanying audio file. This defies our goal of capturing the subtle changes in appearance a human observer would have, nevertheless we provide a comparison in the suppl. material.

\begin{figure}[t]
    \centering
    \includegraphics[width=\textwidth]{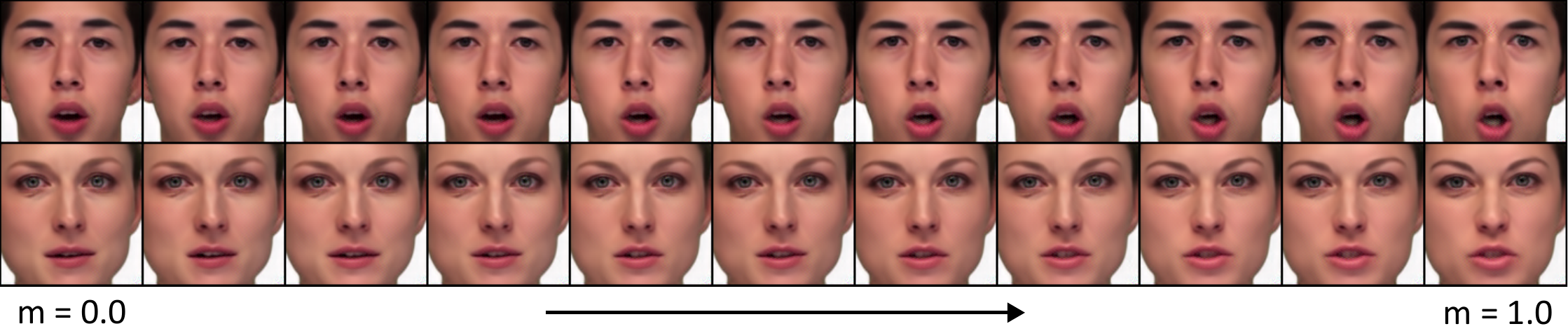}
    \caption{Exploration of modulating the strength of the context offset by increasing \gls{context_weight} (see \cref{sec:methods:eqn:inference_context_weighting}) from $0$ to $1$ from left to right in steps of $0.1$.} \label{sec:experiments:sec:context_augmented_generations:fig:controlling_context_weight}
\end{figure}

\textbf{\textit{How influential is the context in which humans perceive facial expressions?}} 
To empirically evaluate the strength of contextual influence that best captures context effects in humans, we aimed to manually control the computed offset using \gls{context_weight} of \cref{sec:methods:eqn:inference_context_weighting}. As shown in \cref{sec:experiments:sec:context_augmented_generations:fig:controlling_context_weight}, this allowed us to generate facial expressions with varying degrees of contextual influence, which makes them useful as experimental stimuli. The context weight \gls{context_weight} is increased from 0 to 1 in $0.1$ steps from left to right. Note that this weight is independent of the two intensities provided by \gls{ravdess} (\cref{sec:experiments:sec:context_augmented_generations:fig:context_weak_strong}). The smooth transition demonstrates that our model creates continuous representations of the input data.

\subsection{Human Study: Verifying Synthesised Mental Representations}

To assess our model's ability to replicate the contextual impact of emotional speech on facial expression perception in human observers, we conducted two experiments with a total of 160 participants. In \textbf{the first experiment}, 80 participants evaluated neutral facial expressions from the \gls{ravdess} dataset while listening to the depicted actor's speech with either happy or angry prosody. We restricted the study to these two classes because they are of opposite valence. This allowed us to obtain more fine-grained ratings on a continuous Likert scale. We aimed to measure the impact of the audio's emotion on perceived facial expressions. In\textbf{ the second experiment}, a different group of 80 participants rated facial expressions synthesized by our model under happy or angry contextual influences. Each face was presented with five different context weights \gls{context_weight}. 
%We predicted that the difference in ratings between generated happy and angry faces would scale with the intensity level. 
To test whether the model's generations can approximate human-like perception, we compared ratings with those obtained in Experiment 1. This comparison was done for each context weight, determining the parameter that best approximates human responses to contextual influences.

\begin{figure}[t]
    \centering
    \includegraphics[width=0.7\linewidth]{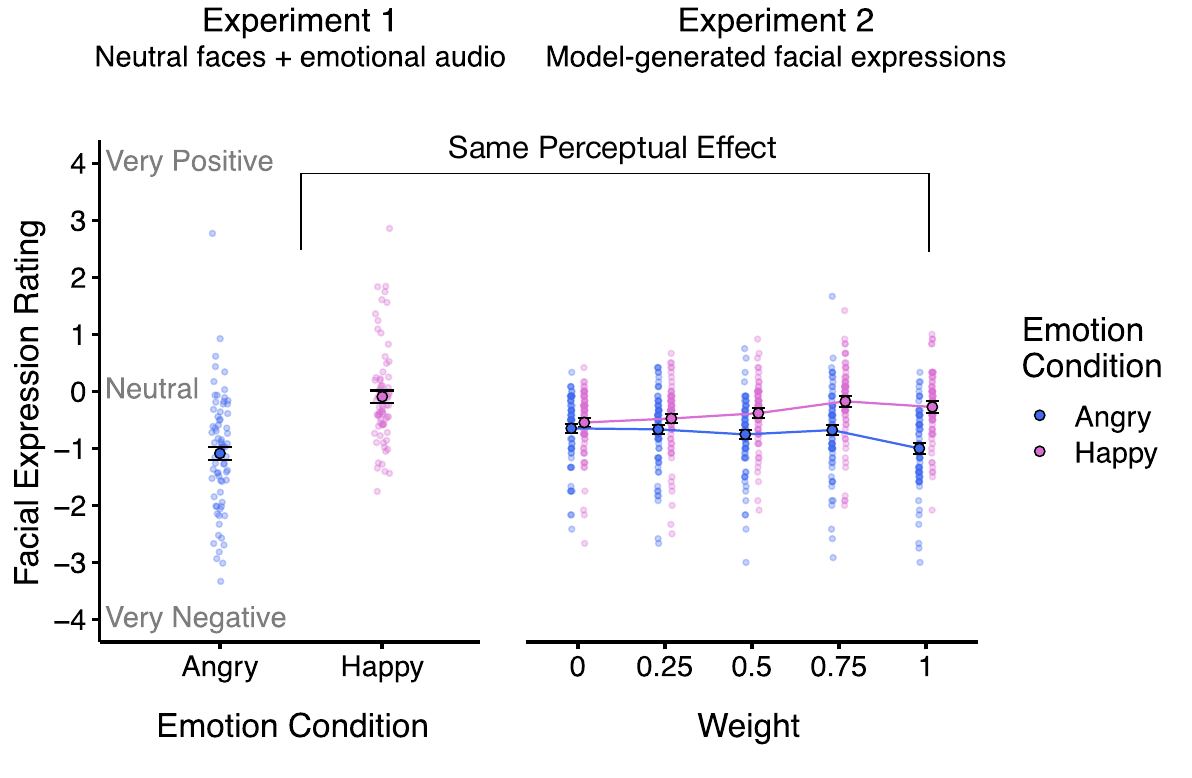}
    \caption{Rating study results depicting mean facial expression ratings in Experiment 1 (left), in which participants rated neutral faces presented in the context of emotional audio, and Experiment 2 (right), in which participants rated faces generated by our model with happy vs. angry audio context (with 5 different values for the context weight \gls{context_weight} of \cref{sec:methods:eqn:inference_context_weighting}). At \gls{context_weight} = 1, ratings are equal across experiments, our model captures the effect of emotional context in human observers. Figure with more statistical details in supplementary material.}\label{sec:experiments:sec:generations_as_approximations:fig:study_results_overview}
\end{figure}

\textbf{Results.} Mean facial expression ratings for different emotional context conditions across both experiments are depicted in \cref{sec:experiments:sec:generations_as_approximations:fig:study_results_overview}. 
In \textbf{experiment 1}, a linear mixed effects model was employed, with the factor emotional context (happy vs. angry audio). A significant effect of emotional context on facial expression ratings was observed ($b = 1.00$, $p < .001$), indicating that identical neutral faces were perceived as more negative when accompanied by angry speech compared to happy speech. For \textbf{experiment 2}, a linear mixed effects model was run with the factors emotional context (happy vs. angry generated expression) and weight (\gls{context_weight} = 0, 0.25, 0.5, 0.75, and 1). To compare rating differences within each weight, the emotional context factor was nested within the weight factor. As shown in  \cref{sec:experiments:sec:generations_as_approximations:fig:study_results_overview} (Experiment 2), expression ratings for faces generated in the context of angry vs. happy prosody did not differ at weights 0 and 0.25 ($bs \leq 0.19$, $ps > .269$), but showed a significant and increasing impact of emotional context at weights 0.5 ($b = 0.37$, $p = .033$), 0.75 ($b = 0.51$, $p = .004$), and 1 ($b = 0.73$, $p < .001$).
\\
\textbf{Discussion.} The results of our rating study reveal two key findings: 1) Our model effectively captures the impact of context, exemplified by emotional prosody, on human facial expression perception. Our generations reflect how a neutral face would subjectively appear to a human observer when associated with a positive or negative context.  2) The model's efficacy in shifting facial appearance towards contextual emotions does not require further adjustment post-training, as evidenced by the optimal context weight being \gls{context_weight} = 1.

\section{Conclusion}

In this work, we introduced a novel approach that simultaneously enhances expression class predictions by taking affective context into account, and provides the means to generate an approximation of the expression a human would perceive. Our model achieves \gls{sota} accuracy on \gls{ravdess} and \gls{mead}, and outperforms joint competitor methods. The implications of our rating study showcase that our model not only accurately quantifies human context-sensitive perception but also successfully mirrors the altered subjective experience back to human observers. This has significant potential, particularly for social artificial agents, that could leverage contextual information to adapt to human mental and emotional states, ensuring successful communication. Our model also addresses the dual nature of context-sensitivity of human perception: on the one hand, leveraging context enhances perceptual efficiency and flexibility \cite{maierKnowledgeaugmentedFacePerception2022, ottenSocialBayesianBrain2017}, while on the other hand, it bears the potential for adversely biased perception, e.g., when contextual information originates from untrustworthy sources \cite{baumClearJudgmentsBased2020}.

\section*{\small{}Acknowledgements}

Funded by the Deutsche Forschungsgemeinschaft (DFG, German Research Foundation) under Germany’s Excellence Strategy – EXC 2002/1 “Science of Intelligence” – project number 390523135. In addition, this work has been partially supported by MIAI@Grenoble Alpes, (ANR-19-P3IA-0003)

\section*{\small{}Disclosure of Interests}

The authors have no competing interests to declare that are relevant to the content of this article.

\bibliographystyle{splncs04}
\bibliography{main}

\clearpage
\setcounter{page}{1}
\maketitlesupplementary

\begin{abstract}
In our work, we presented a model that encodes facial images and audio context using \glspl{vae}. We developed a context fusion network called \acrfull{context_att_net} which shifts the latent facial distribution, to allow more accurate classifications by the classification head, as well as generate an approximation of the facial expression a human would perceive. Here, we provide additional details on the joint training of the \gls{context_att_net} and classification head and provide a derivation of their loss function. We also list more detailed parameters of our rating study. Lastly, we provide more facial expression generations that show the capabilities of our model to purposefully fuse a facial image with affective audio context. The generations are based on neutral face images paired with different audio contexts to mimic the setting in our rating study. In addition, we release the code of our work.
\end{abstract}

\section{Architecture Details}

The number of layers of the components of our model are given in \cref{app:architecture_details}.

\begin{table}
    \centering
    \begin{tabular}{lc}
        \hline
        Component & \# Layers \\ \hline
        VAE Encoder & 16 \\
        VAE Decoder & 16 \\
        \gls{context_att_net} & 3 \\
        Classification Head & 1 \\ \hline
    \end{tabular}
    \caption{Number of layers in each component of our model.}
    \label{app:architecture_details}
\end{table}

\section{Rating Study}

In this section we provide additional information about the procedure used for our rating study with human participants. The study was conducted according to the principles expressed in the Declaration of Helsinki and was approved by the ethics committee of Department of Psychology at Humboldt-Universität zu Berlin. All participants gave their informed consent.

\subsection{Materials}
In \textbf{Experiment 1}, faces of 24 actors from the RAVDESS database with a neutral expression and audio files in which the actors said the sentence “Kids are playing by the door” either with angry or happy emotional prosody served as stimuli. The images were neutral as generated by our network with a weight parameter of 0. We chose a generated face instead of the original frame from the RAVDESS database to eliminate potential effects of low-level visual differences between generated images and images from the database.

In \textbf{Experiment 2}, neutral faces of the same 24 actors were shifted towards the model’s representation of the face in the context of either a happy audio or an angry audio, with five different weights: 0 (i.e. the neutral expression also presented in Experiment 1), 0.25, 0.5, 0.75, and 1 (i.e. the model’s originally trained weight parameter).

In both experiments, we used counterbalancing across participants, such that one participant saw each actor either in the happy or in the angry emotion condition and each face was shown equally as often in each emotion condition.

\subsubsection{Participants}

The study adhered to the principles of the Declaration of Helsinki, approved by the ethics committee of Department of Psychology at Humboldt-Universität zu Berlin. Participants were recruited from Prolific.com and received monetary compensation. For each experiment, 80 participants were recruited. The final samples included 72 English native speakers aged 18–35 years ($M = 29.01$) in Experiment 1 and 77 English native speakers aged 18–35 years ($M = 28.58$) in Experiment 2.
The samples were balanced, with 50\% of participants identifying as female and 50\% identifying as male. We used the following criteria for inclusion in the final samples: Participants who reported not being highly distracted during the experiment, participants who reported not giving random ratings, participants who reported being able to hear all the audio files (Experiment 1), and participants who did not report rating only the audio files (Experiment 1).
Participants were pre-screened for the following criteria based on their data provided to Prolific.com

Age: 18–35;
Prison: No;
Approval Rate: 90–100\%;
Units of alcohol per week: 0, 1-4, or 5-9;
Neurodiversity: No;
Dyslexia: No;
Vision: Yes;
Hearing difficulties: No;
Cochlear implant: No;
Colourblindness: No;
Head Injury—Knock out history: No;
Head Injury: No;
Mental health/illness/condition - ongoing: No;
Medication use: No;
Mild cognitive impairment/Dementia: No;
Autism Spectrum Disorder: No;
Depression: No;
Mental illness daily impact: No;
Anxiety: No;
ADD/ADHD: No;
Anxiety Severity: No;
Mental Health Diagnosis: No;
Mental Health Treatment: None;
First Language: English.

\subsection{Procedure}

Both experiments followed a similar procedure: After providing informed consent, participants rated the facial expressions of stimuli described above in random order on a 9-point Likert scale ranging from “very negative” to “very positive” with “neutral” in the middle. 
In Experiment 1, ratings pertained to a neutral face presented with happy or angry audio, while in Experiment 2, faces modified by our model to depict contextual influence were rated. Post-rating, participants answered questions about their task experience (e.g., whether they were distracted, whether they gave their ratings randomly, and their potential awareness of the hypothesis tested in the study), were debriefed on the study's purpose, and directed back to Prolific.com.

\subsection{Statistical Analyses}

\begin{figure}[t]
    \centering
    \includegraphics[width=0.8\linewidth]{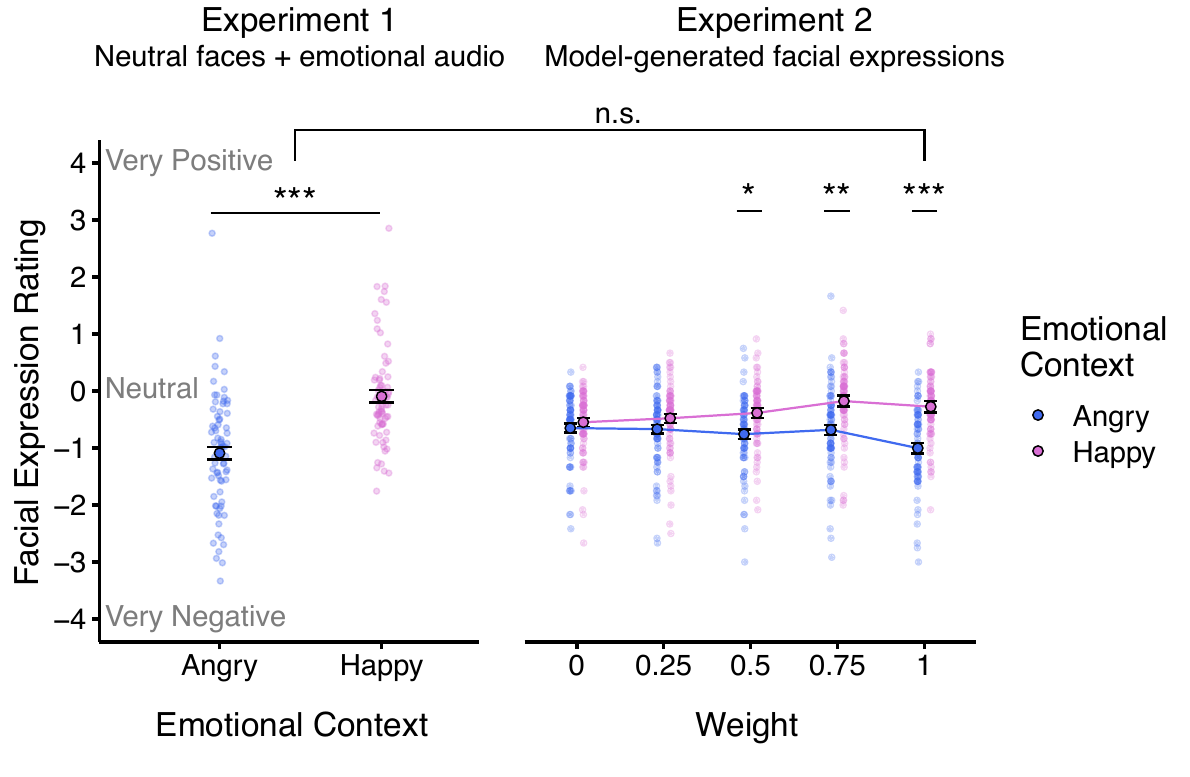}
    \caption{Rating study results \textbf{with additional statistical information}, depicting mean facial expression ratings in Experiment 1 (left), in which participants rated neutral faces presented in the context of emotional audio, and Experiment 2 (right), in which participants rated faces generated by our model with happy vs. angry audio context (with 5 different values for the context weight \gls{context_weight} of \cref{sec:methods:eqn:inference_context_weighting}). At \gls{context_weight} = 1, ratings in Experiment 2 are equal to those of Experiment 1, demonstrating our model's capability to successfully capture the effect of emotional context in human observers. Small dots represent mean ratings per participant, large dots denote grand means across participants, and error bars indicate 95\% confidence intervals. Statistical significance levels are denoted by asterisks: * ($p < .05$), ** ($p < .01$), and *** ($p < .001$);  “n.s.” are non-significant differences.}\label{app:study:results_overview}
\end{figure}

For Experiment 1, we ran a linear mixed effects model with the factor emotion (angry vs. happy audio) coded as a sliding difference contrast (meaning that the estimated effect reflects the predicted mean difference between faces seen with an angry audio vs. a happy audio). 
For Experiment 2, we ran a linear mixed effects model with the factor emotion (generated face shifted towards an angry vs. a happy expression) and the factor weight (with five levels, 0, 0.25, 0.5, 0.75, and 1). To compare the differences in expression ratings between the happy and the angry condition within each of the weights, the factor emotion was nested within the factor weight. 
For both experiments, we modelled random intercepts for participants and items (i.e. depicted actors), as well as random slopes for the effect of emotion over participants and items \cite{batesFittingLinearMixedeffects2015}. The significance of fixed effects coefficients ($p < 0.05$) was tested by Satterthwaite approximation.

\subsection{Results}

Separate models were run for each weight level, including the factors emotional context, experiment (2 vs. 1), and their interaction. With a weight of 1, no significant interaction between experiment and emotional context was found ($b = -0.22$, $p = .084$), whereas this interaction was significant for all other weights ($bs < -0.42$, $ps < .001$). \cref{app:study:results_overview} shows the results including interaction significance.

\section{RAVDESS Comparison to SOTA Generations}\label{app:comparison_sota_generations}

\begin{figure}
    \centering
    \begin{subfigure}[b]{0.7\linewidth}
        \centering
        \begin{tabular}{ccc}
            Ours (Happy) & \citeauthor{sinhaEmotionControllableGeneralizedTalking2022} \cite{sinhaEmotionControllableGeneralizedTalking2022} (Happy) & \citeauthor{maDreamTalkWhenExpressive2023} \cite{maDreamTalkWhenExpressive2023} (Fearful) \\
            \includegraphics[width=0.25\linewidth]{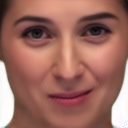} & \includegraphics[width=0.25\linewidth]{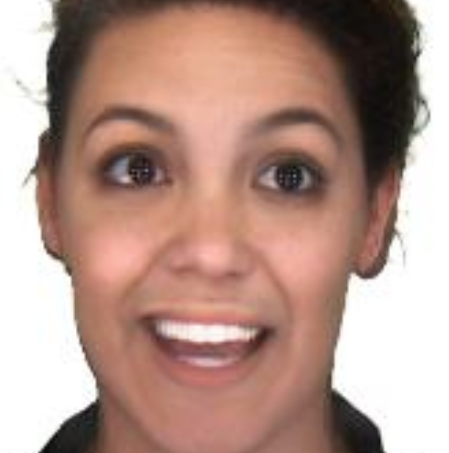} & \includegraphics[width=0.25\linewidth]{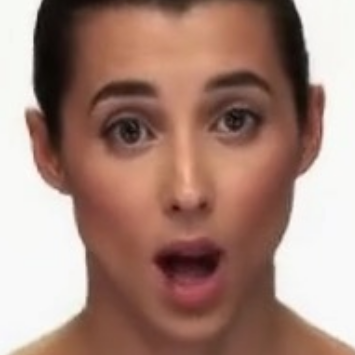} \\
        \end{tabular}
        \caption{Comparison of expression generation with SOTA methods.}
    \end{subfigure}
    \caption{Comparison of our approach to the SOTA generation methods presented in \cite{sinhaEmotionControllableGeneralizedTalking2022,maDreamTalkWhenExpressive2023}. All methods use a neutral input image and generate the result using affective audio context. The two SOTA competitor methods aim at generating a video stream that strongly resembles the emotion in the audio. We aim at synthesizing the subjective facial expression a human observer would perceive.}
    \label{app:comparison_sota_generations:fig:comparison}
\end{figure}

In the main paper we omitted a comparison to SOTA generation methods \cite{sinhaEmotionControllableGeneralizedTalking2022,maDreamTalkWhenExpressive2023,fangFacialExpressionGAN2022} because they pursue a different purpose---making a neutral input images resemble strongly the accompanying audio context. Due to our objective of synthesizing mental representations, such heavy shifts in the facial expression would be too strong for a human observer. Changes in subjective appearance are rather subtle \cite{suessPerceivingEmotionsNeutral2015}. For the sake of completeness, we provide a comparison to SOTA generations in \cref{app:comparison_sota_generations:fig:comparison}. The two SOTA competitors use a target ground-truth sequence they want to model, whereas we compute our adaption only based on the expression class and the model implicilty learns a sensible shift.

\section{Additional Synthesized Facial Expressions}

\begin{figure}[t]
    \centering
    \includegraphics[width=0.6\linewidth]{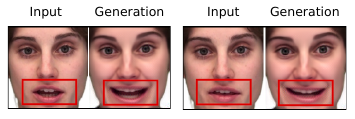}
    \caption{Generations of neutral images paired with happy audio context. The opening of the mouth influences the opening of the mouth generated by our model. Best viewed zoomed in.} \label{app:additional_generations:fig:mouth_opening}
\end{figure}

We provide additional generations of facial expressions on the \gls{ravdess} dataset using our proposed model. \cref{app:additional_generations:fig:mouth_opening} shows that our model picks up the subtle differences in the mouth opening in the generations. \cref{app:additional_generations:fig:context_weight1,app:additional_generations:fig:context_weight2,app:additional_generations:fig:context_weight3} show generations for a neutral input image (left column) paired with different contexts. The context weight \gls{context_weight} is increased from 0 (second column) to 1 (rightmost column) in 0.1 steps. \cref{app:additional_generations:fig:different_neutral_face} shows additional generations for variations of neutral input images with different contexts, to showcase that our model adapts the style of the input image. The leftmost column is the neutral input image, from left to right follows the reconstruction without context, calm audio context, happy audio context, sad audio context, fearful audio context, disgusted audio context, and surprised audio context. \cref{app:additional_generations:fig:normal_strong_audio} shows generations of a neutral input image (top row) paired with normal (middle row) and strong context intensity (bottom row). The \gls{ravdess} dataset provides these two intensities as a binary label for every sample. The context expression classes are from left to right: happy,
sad, angry, fearful, disgusted, surprised.

\def\contextstrengthsize{1.0}

\begin{figure*}[b]
    \centering
    \begin{subfigure}[b]{\contextstrengthsize\textwidth}
        \includegraphics[width=\linewidth]{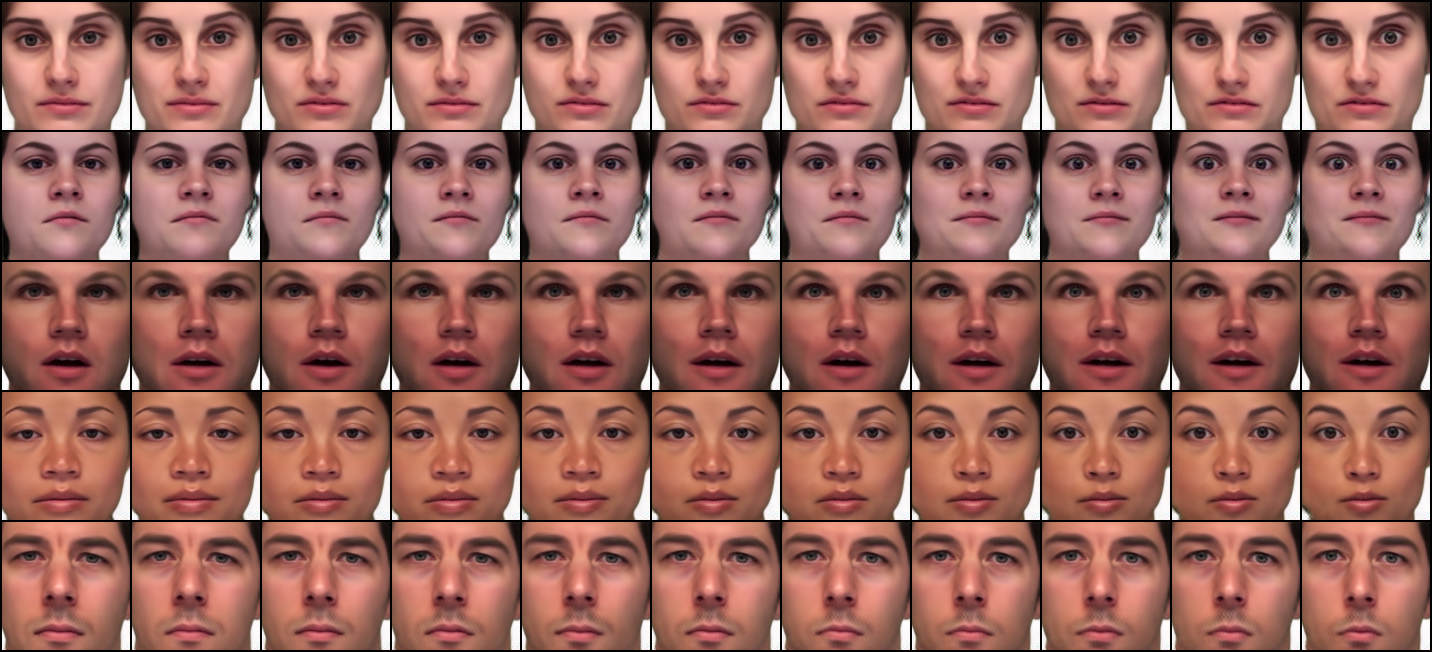}
        \caption{Neutral face with surprised audio context.} 
    \end{subfigure}
    \begin{subfigure}[b]{\contextstrengthsize\textwidth}
        \includegraphics[width=\linewidth]{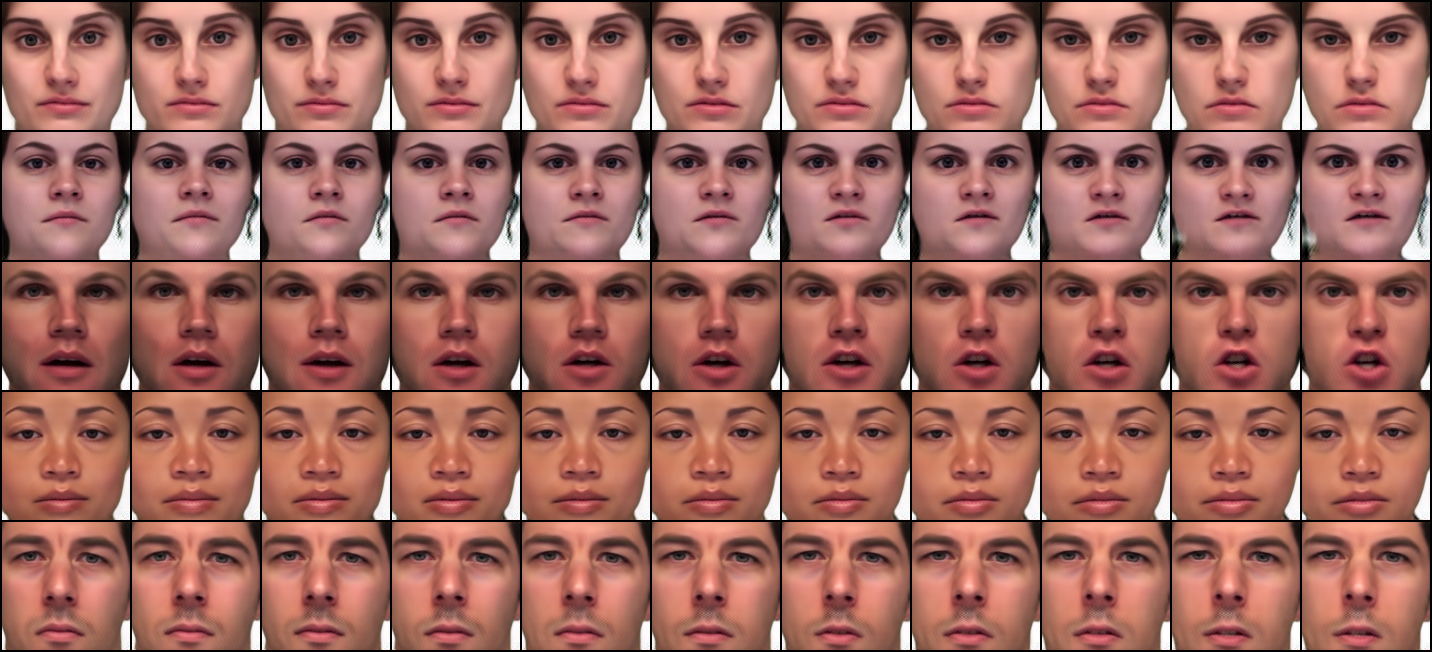}
        \caption{Neutral face with angry audio context.} 
    \end{subfigure}
    \begin{subfigure}[b]{\contextstrengthsize\textwidth}
        \includegraphics[width=\linewidth]{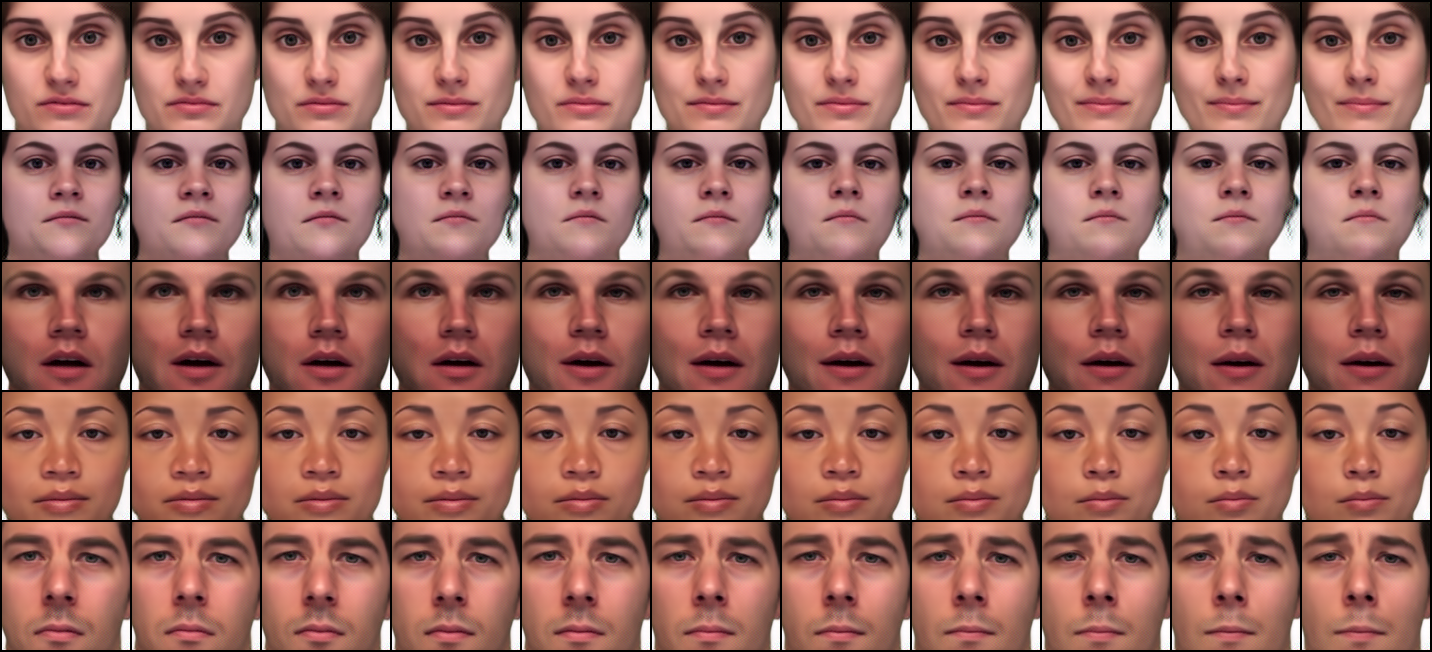}
        \caption{Neutral face with calm audio context.} 
    \end{subfigure}
    \caption{Facial expression generations with varying context weight \gls{context_weight} from 0 to 1 in 0.1 steps.}\label{app:additional_generations:fig:context_weight1}
\end{figure*}

\begin{figure*}
    \centering
    \begin{subfigure}[b]{\contextstrengthsize\textwidth}
        \includegraphics[width=\linewidth]{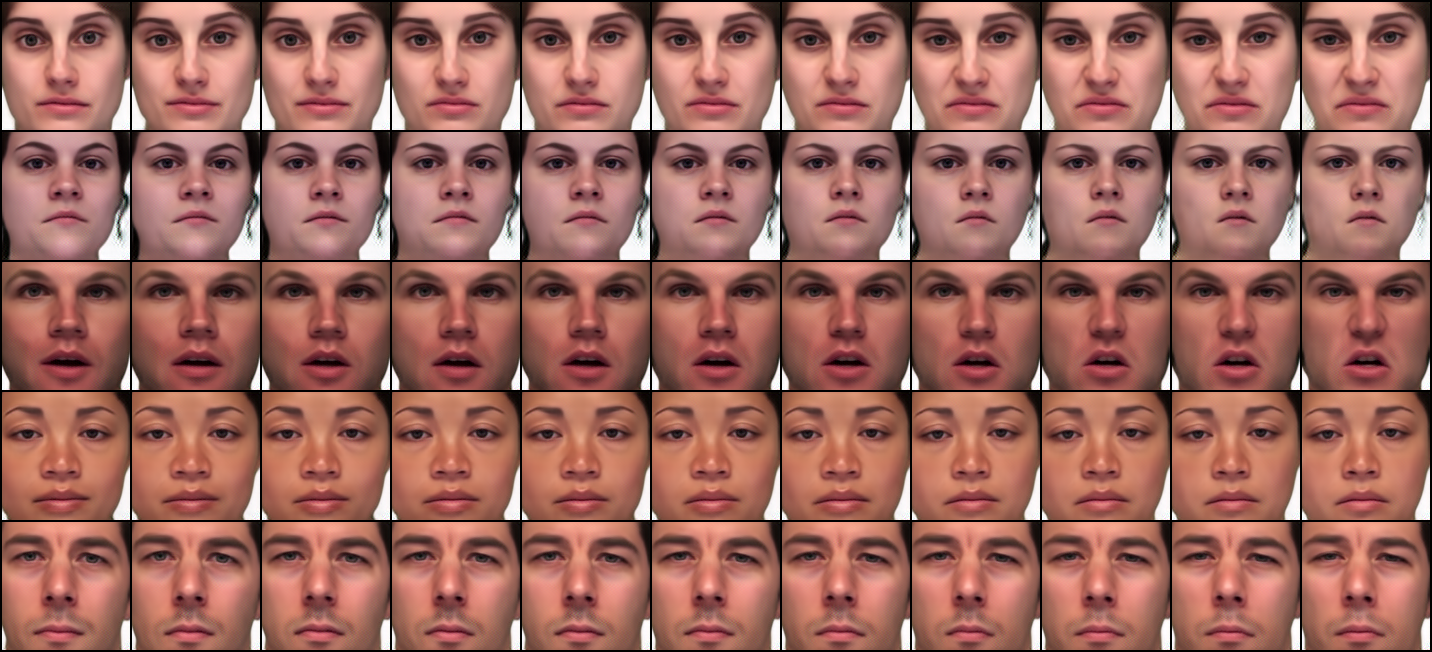}
        \caption{Neutral face with disgusted audio context.} 
    \end{subfigure}
    \begin{subfigure}[b]{\contextstrengthsize\textwidth}
        \includegraphics[width=\linewidth]{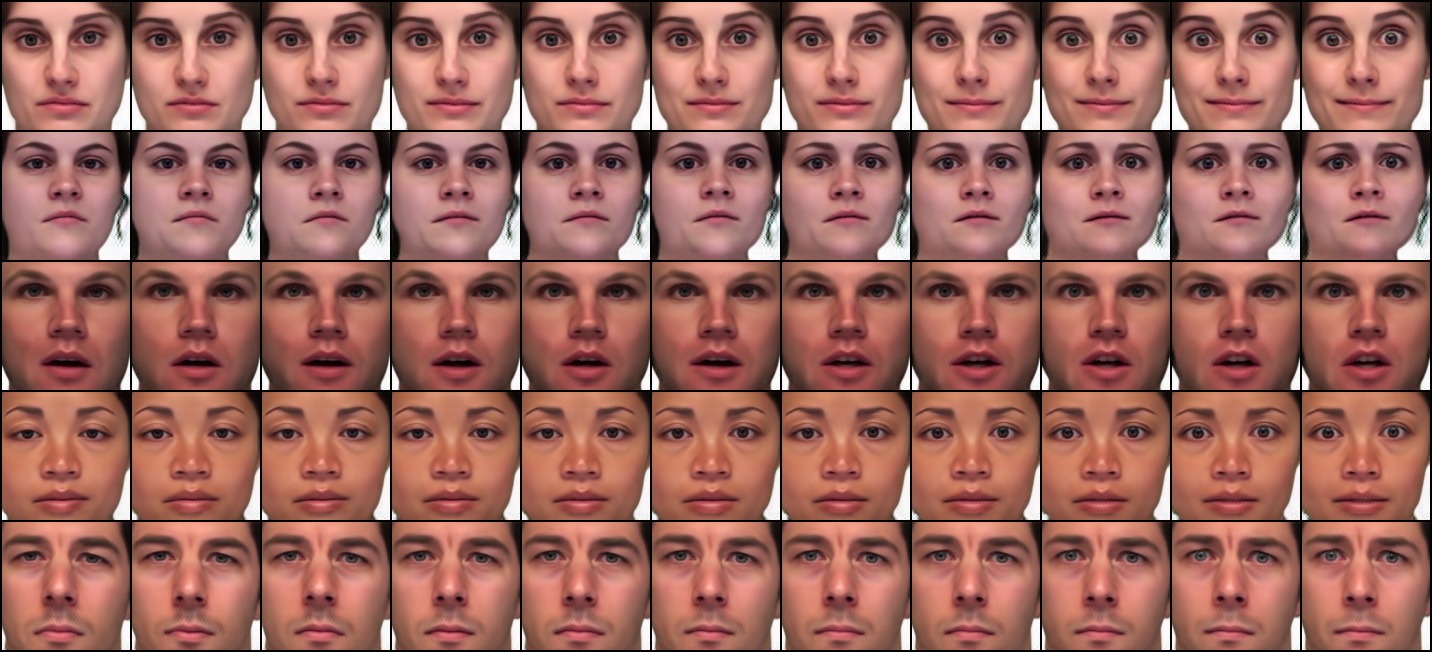}
        \caption{Neutral face with fearful audio context.} 
    \end{subfigure}
    \begin{subfigure}[b]{\contextstrengthsize\textwidth}
        \includegraphics[width=\linewidth]{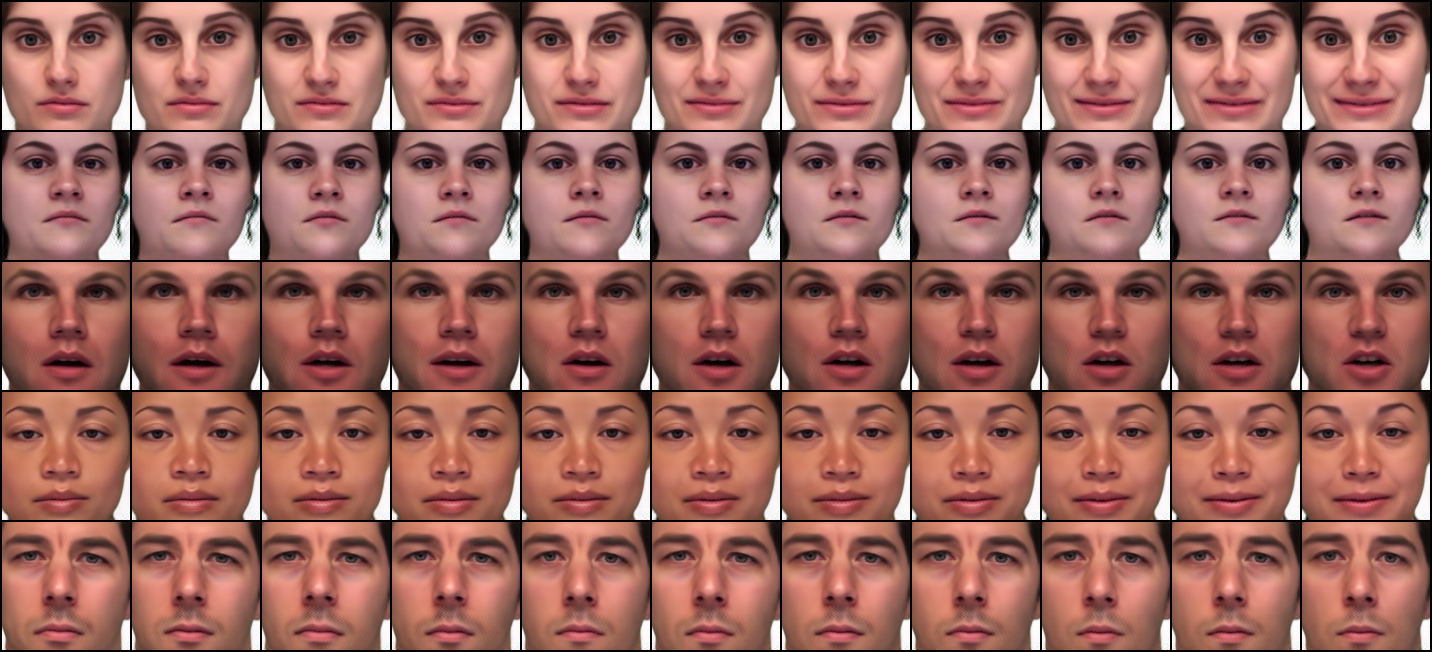}
        \caption{Neutral face with happy audio context.} 
    \end{subfigure}
    \caption{Facial expression generations with varying context weight \gls{context_weight} from 0 to 1 in 0.1 steps (continued).} \label{app:additional_generations:fig:context_weight2}
\end{figure*}

\begin{figure*}
    \centering
    \begin{subfigure}[b]{\contextstrengthsize\textwidth}
        \includegraphics[width=\linewidth]{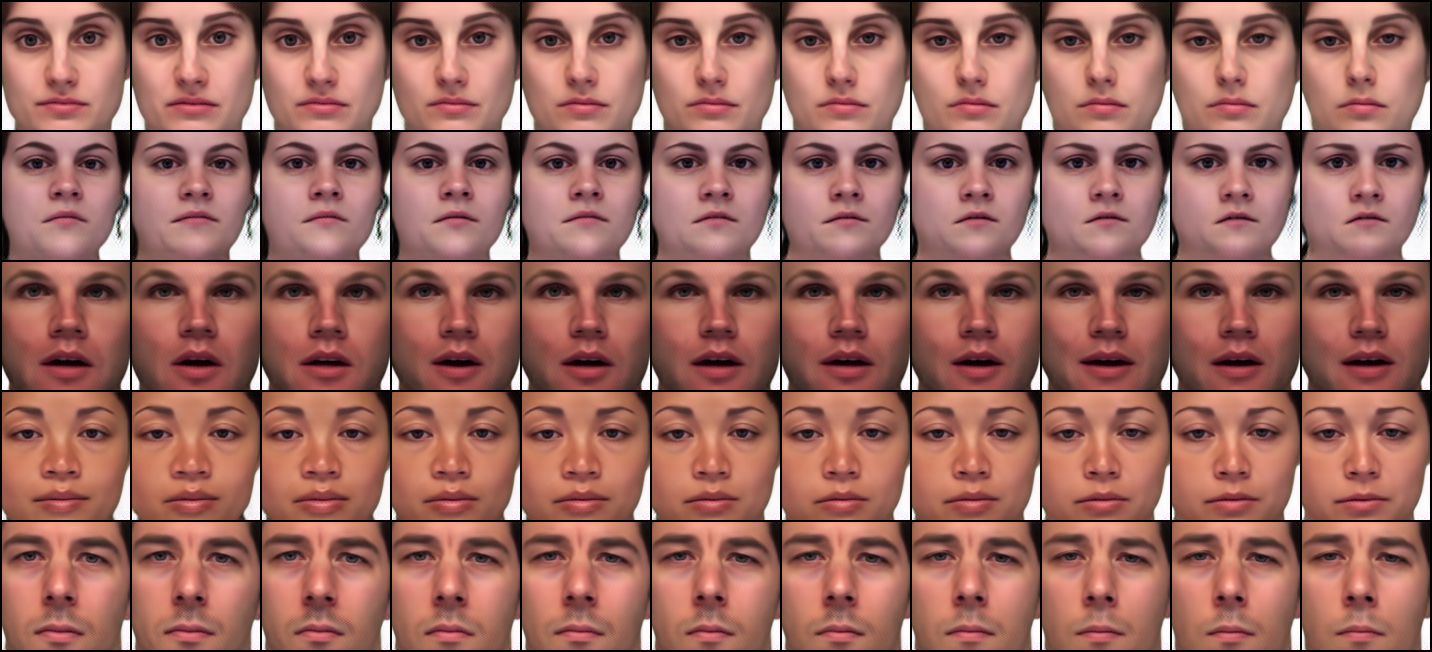}
        \caption{Neutral face with happy audio context.} 
    \end{subfigure}
    \caption{Facial expression generations with varying context weight \gls{context_weight} from 0 to 1 in 0.1 steps (continued).} \label{app:additional_generations:fig:context_weight3}
\end{figure*}

\def\allexpressionsize{0.75}

\begin{figure*}
    \centering
    \begin{subfigure}[b]{\allexpressionsize\textwidth}
        \includegraphics[width=\linewidth]{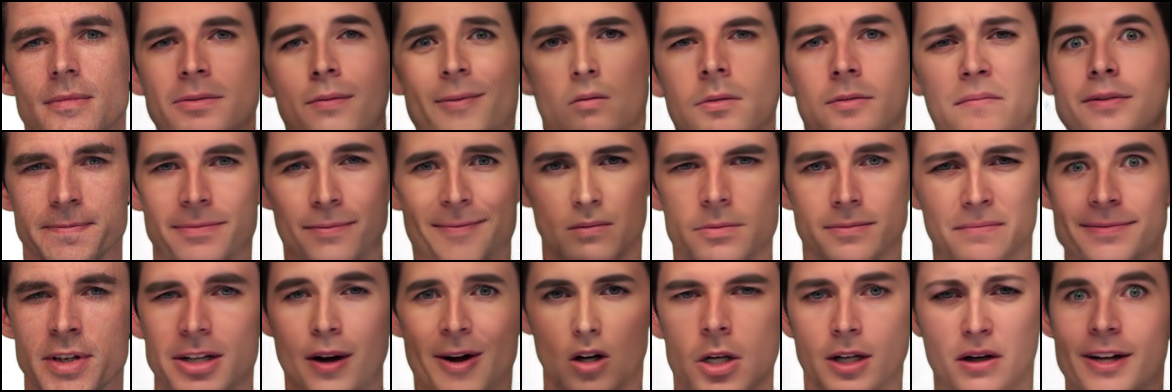}
        \caption{ID 01.} 
    \end{subfigure}
    \begin{subfigure}[b]{\allexpressionsize\textwidth}
        \includegraphics[width=\linewidth]{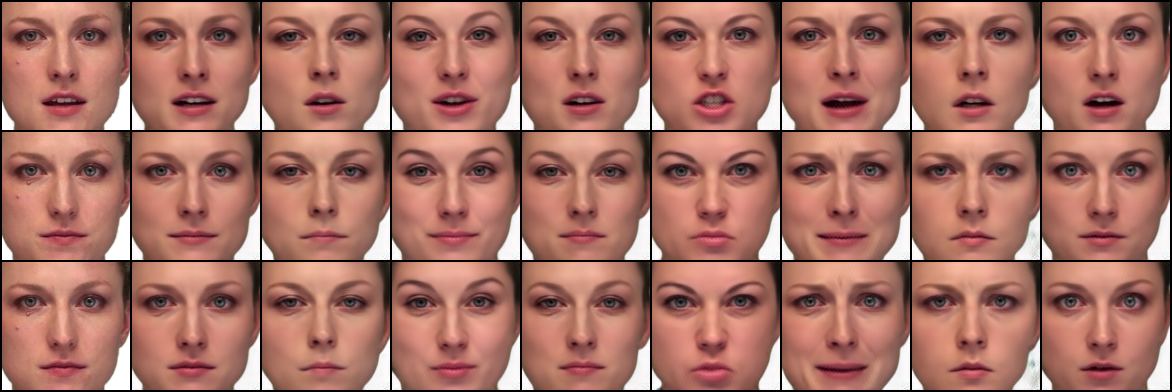}
        \caption{ID 02.} 
    \end{subfigure}
    \begin{subfigure}[b]{\allexpressionsize\textwidth}
        \includegraphics[width=\linewidth]{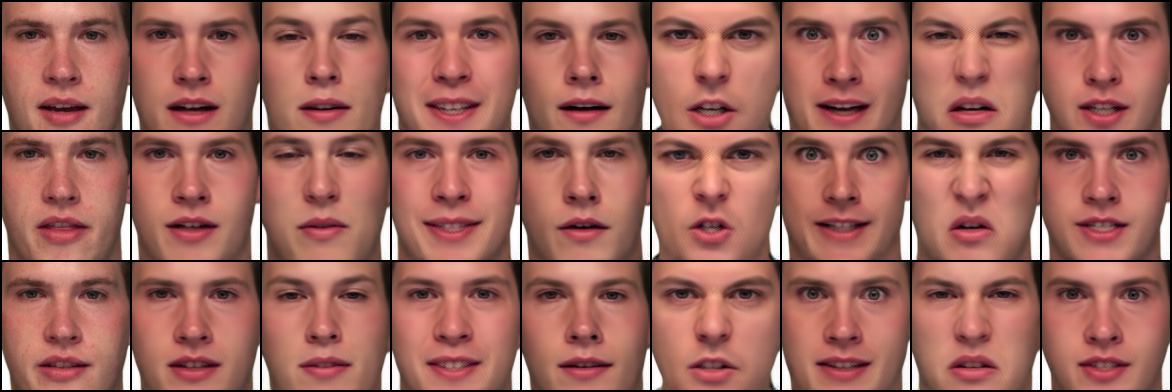}
        \caption{ID 07.} 
    \end{subfigure}
    \begin{subfigure}[b]{\allexpressionsize\textwidth}
        \includegraphics[width=\linewidth]{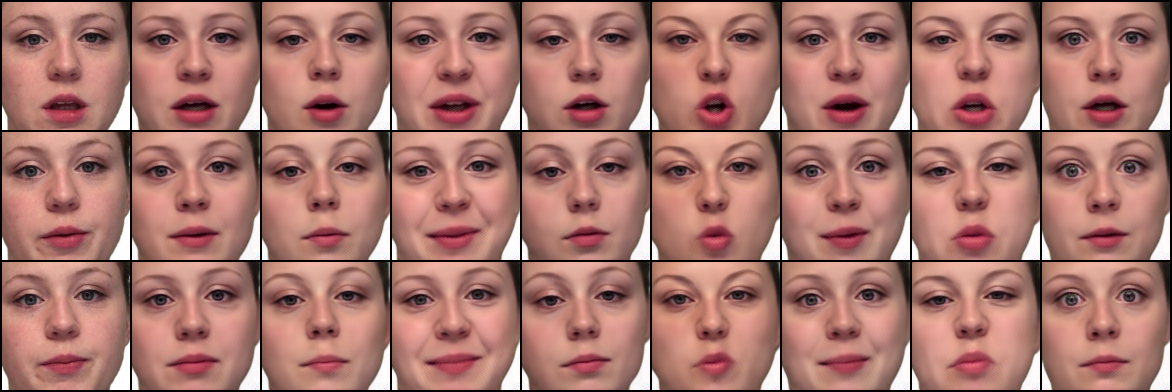}
        \caption{ID 10.} 
    \end{subfigure}
    \begin{subfigure}[b]{\allexpressionsize\textwidth}
        \includegraphics[width=\linewidth]{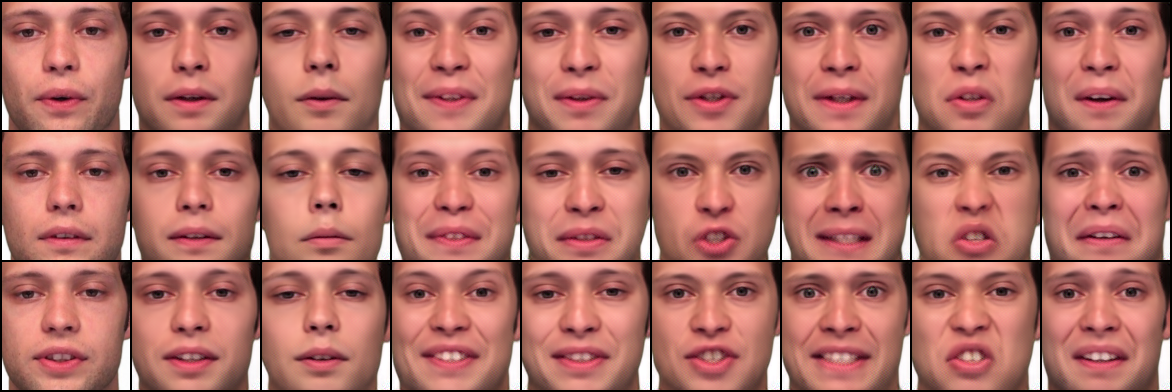}
        \caption{ID 15.} 
    \end{subfigure}
    \caption{Generations of neutral faces paired with contexts of all 7 expressions for a selection of IDs from the \gls{ravdess} dataset. From left to right: Input image, reconstruction without context, calm, happy, sad, angry, fearful, disgusted, surprised. We chose five IDs from \gls{ravdess}.} \label{app:additional_generations:fig:different_neutral_face}
\end{figure*}

\def\weakstrongsize{0.55}

\begin{figure*}
    \centering
    \begin{subfigure}[b]{\weakstrongsize\textwidth}
        \includegraphics[width=\linewidth]{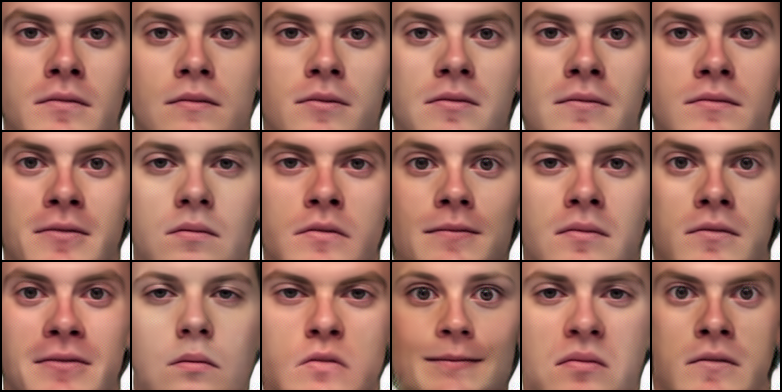}
        \caption{ID 11.} 
    \end{subfigure}
    \begin{subfigure}[b]{\weakstrongsize\textwidth}
        \includegraphics[width=\linewidth]{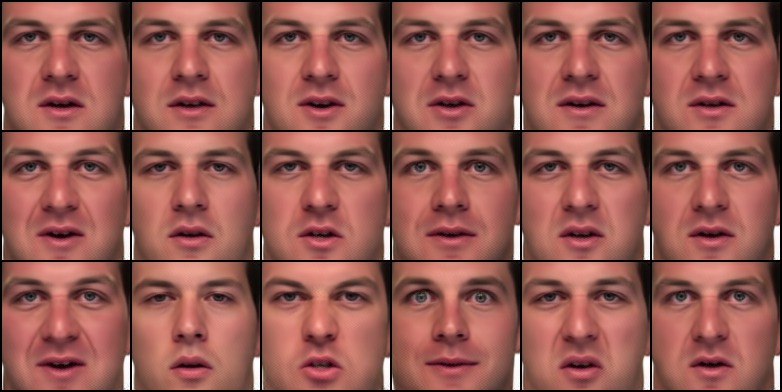}
        \caption{ID 13.} 
    \end{subfigure}
    \begin{subfigure}[b]{\weakstrongsize\textwidth}
        \includegraphics[width=\linewidth]{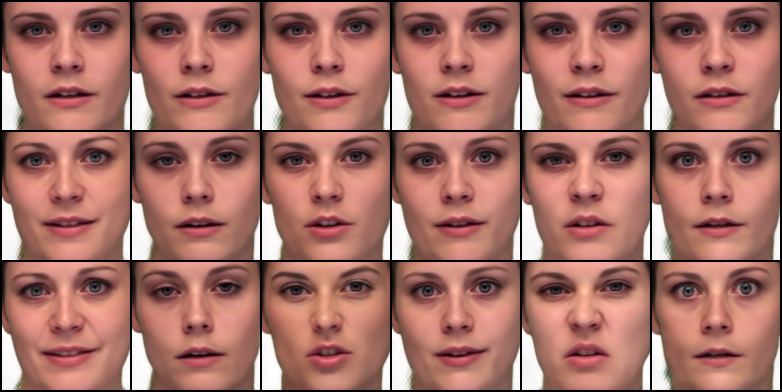}
        \caption{ID 14.} 
    \end{subfigure}
    \caption{Generations of neutral faces paired with contexts of all 7 expressions \textbf{with normal and strong intensity} for a selection of IDs from the \gls{ravdess} dataset. Top row: input image, middle row: normal intensity, bottom row: strong intensity. From left to right: happy, sad, angry, fearful, disgusted, surprised. We chose three IDs from \gls{ravdess}. The \gls{ravdess} comes with \textit{normal} and \textit{strong} binary labels for the expressions.} \label{app:additional_generations:fig:normal_strong_audio}
\end{figure*}

\section{MEAD Generations}

\begin{figure}
    \centering
    \includegraphics[width=\linewidth]{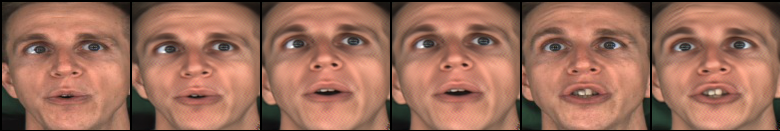}
    \caption{Reconstructions of our method on the \gls{mead} dataset. Pairs of input image (left) and reconstruction (right).}
    \label{app:mead_generations:fig:reconstructions}
\end{figure}

Reconstructions on the \gls{mead} dataset are shown in \cref{app:mead_generations:fig:reconstructions}.

\end{document}